\PassOptionsToPackage{dvipsnames}{xcolor} % must be first
\documentclass[10pt,journal,compsoc]{IEEEtran}
\usepackage{afterpage}
\ifCLASSOPTIONcompsoc
  \usepackage[nocompress]{cite}
\else
  \usepackage{cite}
\fi
\usepackage{graphicx}
\usepackage{amsmath}
\usepackage{array}
\usepackage{fixltx2e}
\usepackage{amssymb}
\usepackage{mathtools}
\usepackage[normalem]{ulem}
\usepackage{epsfig}
\usepackage{multirow}
\usepackage{subfigure} % should probably update to subfig
\usepackage{url}
\usepackage{dirtytalk}
\usepackage{xcolor}
\newcommand{\FF}{\mathbf{F}}
\newcommand{\XX}{\mathbf{X}}
\newcommand{\RR}{\mathbb{R}}
\newcommand{\UU}{\mathbf{U}}
\newcommand{\VV}{\mathbf{V}}
\newcommand{\WW}{\mathbf{W}}
\newcommand{\uu}{\mathbf{u}}
\newcommand{\vv}{\mathbf{v}}
\newcommand{\zz}{\mathbf{z}}
\newcommand{\xx}{\mathbf{x}}

\begin{document}
\bstctlcite{IEEEexample:BSTcontrol}
\title{Squeeze-and-Excitation Networks}

\author{Jie Hu$^{[0000-0002-5150-1003]}$ \quad% <-this % stops a space
\IEEEcompsocitemizethanks{\IEEEcompsocthanksitem Jie Hu and Enhua Wu are with the State Key Laboratory of Computer Science, Institute of Software, Chinese Academy of Sciences, Beijing, 100190, China. \protect\\
They are also with the University of Chinese Academy of Sciences, Beijing, 100049, China. \protect\\ 
Jie Hu is also with Momenta and Enhua Wu is also with the Faculty of Science and Technology \& AI Center at University of Macau. \protect\\
E-mail: \texttt{hujie@ios.ac.cn} \quad \texttt{ehwu@umac.mo}

\IEEEcompsocthanksitem Gang Sun is with LIAMA-NLPR at the Institute of Automation, Chinese Academy of Sciences. He is also with Momenta. \protect\\
E-mail: \texttt{sungang@momenta.ai}

\IEEEcompsocthanksitem Li Shen and Samuel Albanie are with the Visual Geometry Group at the University of Oxford.\protect\\
E-mail: \texttt{\{lishen,albanie\}@robots.ox.ac.uk}
}% <-this % stops an unwanted space
Li Shen$^{[0000-0002-2283-4976]}$ \quad
Samuel Albanie$^{[0000-0001-9736-5134]}$ \quad \\
Gang Sun$^{[0000-0001-6913-6799]}$ \quad
Enhua Wu$^{[0000-0002-2174-1428]}$
}

\IEEEtitleabstractindextext{%
\begin{abstract}

The central building block of convolutional neural networks (CNNs) is the convolution operator, which enables networks to construct informative features by fusing both spatial and channel-wise information within local receptive fields at each layer. A broad range of prior research has investigated the spatial component of this relationship, seeking to strengthen the representational power of a CNN by enhancing the quality of spatial encodings throughout its feature hierarchy. In this work, we focus instead on the channel relationship and propose a novel architectural unit, which we term the ``Squeeze-and-Excitation" (SE) block, that adaptively recalibrates channel-wise feature responses by explicitly modelling interdependencies between channels.  We show that these blocks can be stacked together to form SENet architectures that generalise extremely effectively across different datasets. We further demonstrate that SE blocks bring significant improvements in performance for existing state-of-the-art CNNs at slight additional computational cost. Squeeze-and-Excitation Networks formed the foundation of our ILSVRC $2017$ classification submission which won first place and reduced the top-$5$ error to $2.251\%$, surpassing the winning entry of $2016$ by a relative improvement of ${\sim}25\%$.  Models and code are available at \url{https://github.com/hujie-frank/SENet}.
\end{abstract}

}

\maketitle

\IEEEdisplaynontitleabstractindextext

% \fi
\IEEEpeerreviewmaketitle

%\IEEEpubidadjcol

\IEEEraisesectionheading{\section{Introduction}\label{sec:introduction}}

\IEEEPARstart{C}{onvolutional} neural networks (CNNs) have proven to be useful models for tackling a wide range of visual tasks \cite{krizhevsky_nips2012alex, toshev_cvpr2014pose, long_cvpr2015fcn, ren_nips2015fasterrcnn}. At each convolutional layer in the network, a collection of filters expresses neighbourhood spatial connectivity patterns along input channels---fusing spatial and channel-wise information together within local receptive fields.  By interleaving a series of convolutional layers with non-linear activation functions and downsampling operators, CNNs are able to produce image representations that capture hierarchical patterns and attain global theoretical receptive fields.
A central theme of computer vision research is the search for more powerful representations that capture only those properties of an image that are most salient for a given task, enabling improved performance.  As a widely-used family of models for vision tasks, the development of new neural network architecture designs now represents a key frontier in this search.
Recent research has shown that the representations produced by CNNs can be strengthened by integrating learning mechanisms into the network that help capture spatial correlations between features.  
One such approach, popularised by the Inception family of architectures \cite{szegedy_cvpr2015googlenet, ioffe_icml2015bn}, incorporates multi-scale processes into network modules to achieve improved performance. Further work has sought to better model spatial dependencies \cite{bell_cvpr2016ionet, newell_eccv2016hourglass} and incorporate spatial attention into the structure of the network \cite{max_nips2015stn}.

%--------------------------------------------------------------------------
\begin{figure*}[t]
\begin{center}
\includegraphics[width=0.85\textwidth]{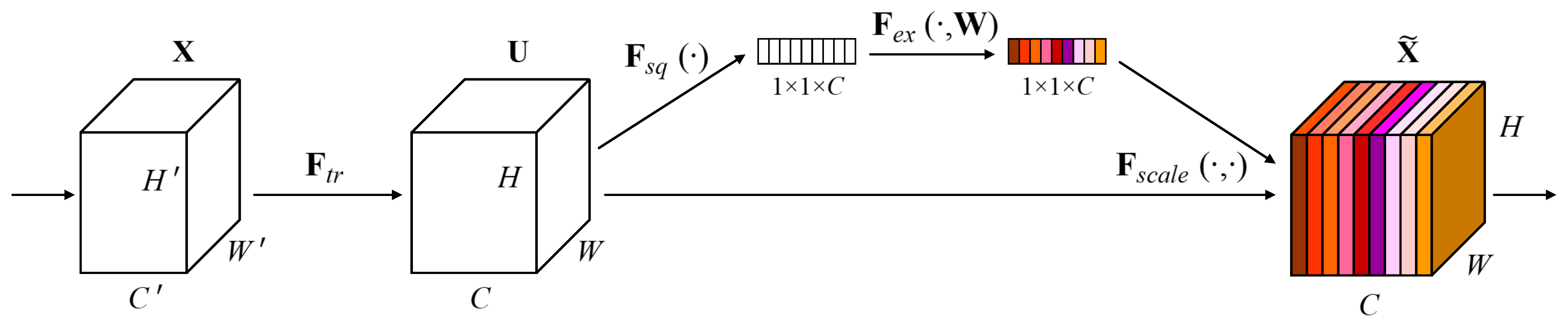}
\end{center}
\caption{A Squeeze-and-Excitation block.}
\label{fig:splash}
\end{figure*}

In this paper, we investigate a different aspect of network design - the relationship between channels.  We introduce a new architectural unit, which we term the {\it Squeeze-and-Excitation} (SE) block, with the goal of improving the quality of representations produced by a network by explicitly modelling the interdependencies between the channels of its convolutional features. To this end, we propose a mechanism that allows the network to perform feature recalibration, through which it can learn to use global information to selectively emphasise informative features and suppress less useful ones.

The structure of the SE building block is depicted in Fig.~\ref{fig:splash}.  For any given transformation $\FF_{tr}$ mapping the input $\XX$ to the feature maps $\UU$ where $\UU\in \mathbb{R}^{H \times W \times C}$, e.g. a convolution, we can construct a corresponding SE block to perform feature recalibration. 
The features $\UU$ are first passed through a \textit{squeeze} operation, which produces a channel descriptor by aggregating feature maps across their spatial dimensions ($H \times W$). The function of this descriptor is to produce an embedding of the global distribution of channel-wise feature responses, allowing information from the global receptive field of the network to be used by all its layers.  The aggregation is followed by an \textit{excitation} operation, which takes the form of a simple self-gating mechanism that takes the embedding as input and produces a collection of per-channel modulation weights. These weights are applied to the feature maps $\UU$ to generate the output of the SE block which can be fed directly into subsequent layers of the network.

It is possible to construct an SE network (SENet) by simply stacking a collection of SE blocks. Moreover, these SE blocks can also be used as a drop-in replacement for the original block at a range of depths in the network architecture (Section~\ref{subsec:ablation-stages}). While the template for the building block is generic, the role it performs at different depths differs throughout the network.  In earlier layers, it excites informative features in a class-agnostic manner, strengthening the shared low-level representations.  In later layers, the SE blocks become increasingly specialised, and respond to different inputs in a highly class-specific manner (Section~\ref{subsection:role-of-excitation}).  As a consequence, the benefits of the feature recalibration performed by SE blocks can be accumulated through the network.

The design and development of new CNN architectures is a difficult engineering task, typically requiring the selection of many new hyperparameters and layer configurations.  By contrast, the structure of the SE block is simple and can be used directly in existing state-of-the-art architectures by replacing components with their SE counterparts, where the performance can be effectively enhanced. SE blocks are also computationally lightweight and impose only a slight increase in model complexity and computational burden. 

To provide evidence for these claims, we develop several SENets and conduct an extensive evaluation on the ImageNet dataset \cite{russakovsky_ijcv2015imagenet}. We also present results beyond ImageNet that indicate that the benefits of our approach are not restricted to a specific dataset or task.
By making use of SENets, we ranked first in the ILSVRC 2017 classification competition. Our best model ensemble achieves a $2.251\%$ top-5 error on the test set\footnote{\scriptsize{\url{http://image-net.org/challenges/LSVRC/2017/results}}}. This represents roughly a $25\%$ relative improvement when compared to the winner entry of the previous year (top-5 error of $2.991\%$).
\section{Related Work\label{sec:related}}

\noindent\textbf{Deeper architectures.} %
VGGNets~\cite{simonyan_iclr2015vgg} and Inception models~\cite{szegedy_cvpr2015googlenet} showed that increasing the depth of a network could significantly increase the quality of representations that it was capable of learning. By regulating the distribution of the inputs to each layer, Batch Normalization (BN)~\cite{ioffe_icml2015bn} added stability to the learning process in deep networks and produced smoother optimisation surfaces~\cite{santurkar_nips2018}. Building on these works, ResNets demonstrated that it was possible to learn considerably deeper and stronger networks through the use of identity-based skip connections~\cite{he_cvpr2016resnet, he_eccv2016preact}. Highway networks \cite{srivastava_nips2015highway} introduced a gating mechanism to regulate the flow of information along shortcut connections. Following these works, there have been further reformulations of the connections between network layers \cite{chen_nips2017dpn, huang_cvpr2017dns}, which show promising improvements to the learning and representational properties of deep networks.

An alternative, but closely related line of research has focused on methods to improve the functional form of the computational elements contained within a network.  Grouped convolutions have proven to be a popular approach for increasing the cardinality of learned transformations \cite{ioannou_cvpr2017roots, xie_cvpr2017resnext}. More flexible compositions of operators can be achieved with multi-branch convolutions \cite{ioffe_icml2015bn, szegedy_cvpr2015googlenet, szegedy_cvpr2016inceptionv3, szegedy_iclrw2016inceptionv4}, which can be viewed as a natural extension of the grouping operator. In prior work, cross-channel correlations are typically mapped as new combinations of features, either independently of spatial structure \cite{max_bmvc2014lowrank, chollet_cvpr2017xception} or jointly by using standard convolutional filters \cite{lin_iclr2014nin} with $1\times1$ convolutions. Much of this research has concentrated on the objective of reducing model and computational complexity, reflecting an assumption that channel relationships can be formulated as a composition of instance-agnostic functions with local receptive fields.  In contrast, we claim that providing the unit with a mechanism to explicitly model dynamic, non-linear dependencies between channels using global information can ease the learning process, and significantly enhance the representational power of the network.

\vspace{5pt}
\noindent\textbf{Algorithmic Architecture Search.} Alongside the works described above, there is also a rich history of research that aims to forgo manual architecture design and instead seeks to learn the structure of the network automatically.  Much of the early work in this domain was conducted in the neuro-evolution community, which established methods for searching across network topologies with evolutionary methods \cite{miller_1989,stanley_2002}. While often computationally demanding, evolutionary search has had notable successes which include finding good memory cells for sequence models \cite{bayer_icann2009,jozefowicz_icml2015} and learning sophisticated architectures for large-scale image classification \cite{xie_iccv2017,real_icml2017,real_2018a}.  With the goal of reducing the computational burden of these methods, efficient alternatives to this approach have been proposed based on Lamarckian inheritance~\cite{elsken_2018mul} and differentiable architecture search~\cite{liu_2018darts}.

By formulating architecture search as hyperparameter optimisation, random search \cite{bergstra_jmlr2012} and other more sophisticated model-based optimisation techniques \cite{liu_eccv2018progressive,negrinho_2017deeparchitect} can also be used to tackle the problem.  Topology selection as a path through a fabric of possible designs \cite{saxena_nips2016} and direct architecture prediction \cite{brock_iclr2018,baker_iclrw2018} have been proposed as additional viable architecture search tools. Particularly strong results have been achieved with techniques from reinforcement learning \cite{baker_iclr2017,zoph_iclr2017,zoph_cvpr2018,liu_iclr2018deepmind,pham_icml2018}.  SE blocks can be used as atomic building blocks for these search algorithms, and were demonstrated to be highly effective in this capacity in concurrent work \cite{tan_2018mnasnet}.

\vspace{5pt}
\noindent\textbf{Attention and gating mechanisms.}  Attention can be interpreted as a means of biasing the allocation of available computational resources towards the most informative components of a signal \cite{olshausen_1993, itti_1998model, itti_2001, larochelle_nips2010, mnih_nips2014, vaswani_nips2017}.  Attention mechanisms have demonstrated their utility across many tasks including sequence learning \cite{bluche_nips2016text, miech_2017contextgate}, localisation and understanding in images \cite{cao_iccv2015twice, max_nips2015stn}, image captioning \cite{xu_icml2015, chen_cvpr2017sca} and lip reading \cite{chung_cvpr2017lip}. In these applications, it can  be incorporated as an operator following one or more layers representing higher-level abstractions for adaptation between modalities.   Some works provide interesting studies into the combined use of spatial and channel attention \cite{wangfei_cvpr2017resattent, woo_eccv2018cbam}. Wang et al. \cite{wangfei_cvpr2017resattent} introduced a powerful trunk-and-mask attention mechanism based on hourglass modules \cite{newell_eccv2016hourglass} that is inserted between the intermediate stages of deep residual networks. By contrast, our proposed SE block comprises a lightweight gating mechanism which focuses on enhancing the representational power of the network by modelling channel-wise relationships in a computationally efficient manner.

\section{Squeeze-and-Excitation Blocks\label{sec:se-blocks}}

A Squeeze-and-Excitation block is a computational
unit which can be built upon a transformation $\FF_{tr}$ mapping an input $\XX \in \RR^{H' \times W' \times C'}$ to feature maps $\UU \in \RR^{H \times W \times C}$.  In the notation that follows we take $\FF_{tr}$ to be a convolutional operator and use $\VV= [\vv_1, \vv_2, \dots, \vv_{C}]$ to denote the learned set of filter kernels, where $\vv_c$ refers to the parameters of the \mbox{$c$-th} filter.  We can then write the outputs as $\UU = [\uu_1, \uu_2, \dots, \uu_{C}]$, where

\begin{equation}\label{simple_neuron}
\uu_c = \vv_c \ast \XX = \sum_{s=1}^{C'}\vv^s_c \ast \xx^s.
\end{equation}
Here $\ast$ denotes convolution, $\vv_c = [\vv^1_c, \vv^2_c, \dots, \vv^{C'}_c]$, $\XX = [\xx^1, \xx^2, \dots, \xx^{C'}]$ and \mbox{$\uu_c \in \RR^{H \times W}$}. $\vv^s_c$ is a $2$D spatial kernel representing a single channel of $\vv_c$ that acts on the corresponding channel of $\XX$. To simplify the notation, bias terms are omitted. Since the output is produced by a summation through all channels, channel dependencies are implicitly embedded in $\vv_c$, but are entangled with the local spatial correlation captured by the filters.  The channel relationships modelled by convolution are inherently implicit and local (except the ones at top-most layers). We expect the learning of convolutional features to be enhanced by explicitly modelling channel interdependencies, so that the network is able to increase its sensitivity to informative features which can be exploited by subsequent transformations. Consequently, we would like to provide it with access to global information and recalibrate filter responses in two steps, \textit{squeeze} and \textit{excitation}, before they are fed into the next transformation. A diagram illustrating the structure of an SE block is shown in Fig.~\ref{fig:splash}.

\subsection{Squeeze: Global Information Embedding}
In order to tackle the issue of exploiting channel dependencies, we first consider the signal to each channel in the output features.  Each of the learned filters operates with a local receptive field and consequently each unit of the transformation output $\UU$ is unable to exploit contextual information outside of this region.

To mitigate this problem, we propose to {\it squeeze} global spatial information into a channel descriptor. This is achieved by using global average pooling to generate channel-wise statistics. Formally, a statistic $\zz \in \RR^{C}$ is generated by shrinking $\UU$ through its spatial dimensions $H \times W$, such that the $c$-th element of $\zz$ is calculated by:
%\vspace{-1mm}
\begin{equation}\label{squeeze}
z_c = \FF_{sq}(\uu_c) = \frac{1}{H\times W}\sum_{i=1}^{H} \sum_{j=1}^{W} u_c(i,j).
\end{equation}

{\it Discussion.} The output of the transformation $\UU$ can be interpreted as a collection of the local descriptors whose statistics are expressive for the whole image. Exploiting such information is prevalent in prior feature engineering work \cite{yang_cvpr2009spm, fisher_ijcv2013fv, shen_tip2015}. We opt for the simplest aggregation technique, global average pooling, noting that more sophisticated strategies could be employed here as well.

%-------------------------------------------------------------------------
\subsection{Excitation: Adaptive Recalibration} \label{subsec:adaptive-recal}
To make use of the information aggregated in the \textit{squeeze} operation, we follow it with a second operation which aims to fully capture channel-wise dependencies.  To fulfil this objective, the function must meet two criteria: first, it must be flexible (in particular, it must be capable of learning a nonlinear interaction between channels) and second, it must learn a non-mutually-exclusive relationship since we would like to ensure that multiple channels are allowed to be emphasised (rather than enforcing a one-hot activation). To meet these criteria, we opt to employ a simple gating mechanism with a sigmoid activation:

\begin{equation}\label{excitation}
\mathbf{s} = \mathbf{F}_{ex}(\mathbf{z}, \mathbf{W}) = \sigma(g(\mathbf{z}, \mathbf{W})) = \sigma(\WW_2\delta(\WW_1\mathbf{z})),
\end{equation}
where $\delta$ refers to the ReLU \cite{nair_icml2010relu} function, $\WW_1 \in \RR^{\frac{C}{r} \times C}$ and $\WW_2 \in \RR^{C \times \frac{C}{r}}$. To limit model complexity and aid generalisation, we parameterise the gating mechanism by forming a bottleneck with two fully-connected (FC) layers around the non-linearity, i.e. a dimensionality-reduction layer with reduction ratio $r$ (this parameter choice is discussed in Section~\ref{subsec:reduction}), a ReLU and then a dimensionality-increasing layer returning to the channel dimension of the transformation output $\UU$. The final output of the block is obtained by rescaling $\UU$ with the activations $\mathbf{s}$:

\begin{equation}\label{scale}
\widetilde{\mathbf{x}}_c = \mathbf{F}_{scale}(\mathbf{u}_c, s_c) = s_c\,\mathbf{u}_c,
\end{equation}
where $\widetilde{\XX} = [\widetilde{\xx}_1, \widetilde{\xx}_2, \dots, \widetilde{\xx}_{C}]$ and $\FF_{scale}(\uu_c, s_c)$ refers to channel-wise multiplication between the scalar $s_c$ and the feature map $\uu_c \in \RR^{H \times W}$.

{\it Discussion.}  The excitation operator maps the input-specific descriptor $\mathbf{z}$ to a set of channel weights. In this regard, SE blocks intrinsically introduce dynamics conditioned on the input, which can be regarded as a self-attention function on channels whose relationships are not confined to the local receptive field the convolutional filters are responsive to.
\subsection{Instantiations}
\label{se_inception_resnet}

\begin{figure}[t]
\begin{center}
\includegraphics[height=65mm]{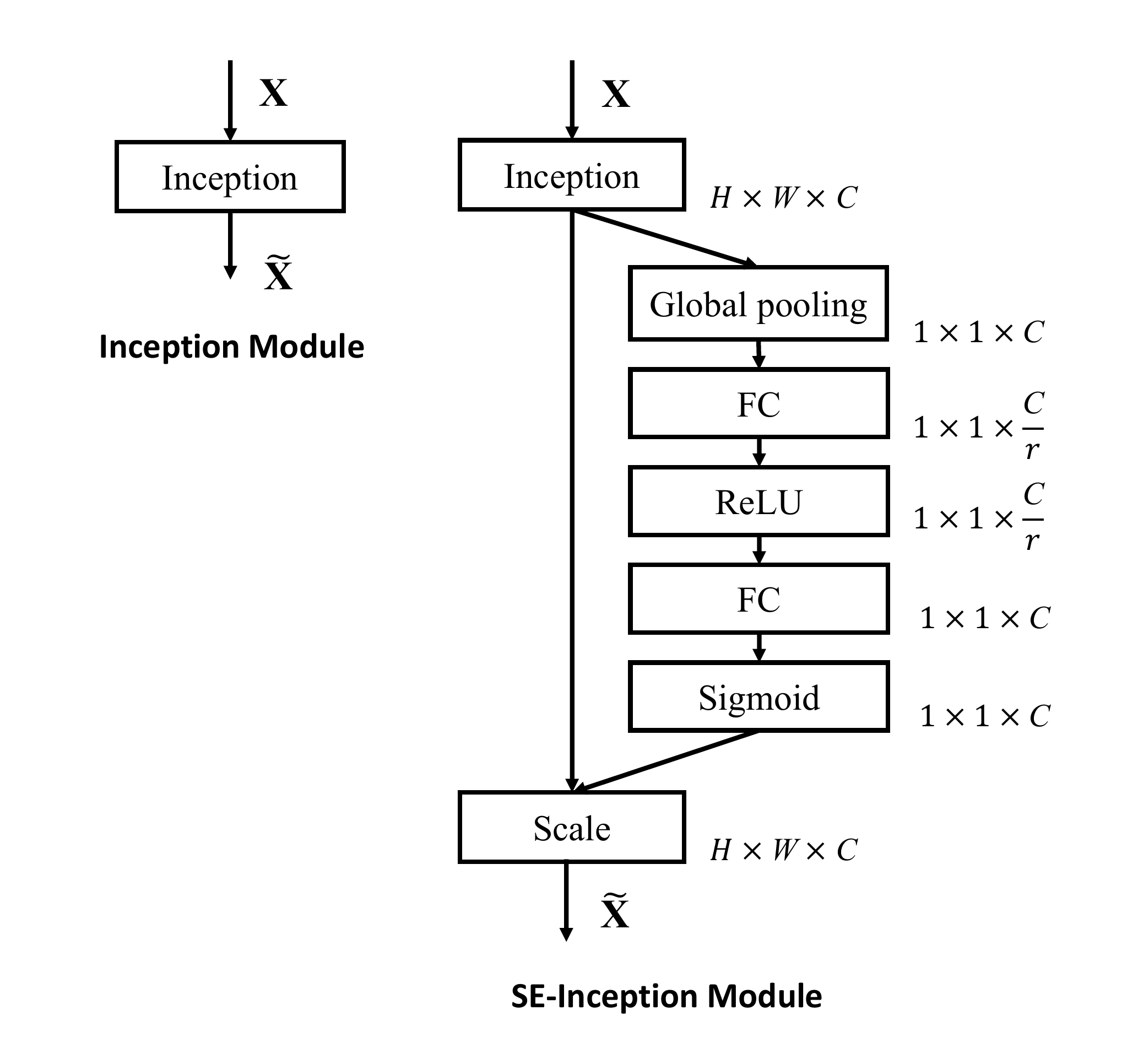}
\end{center}
\vspace{-2em}
\caption{The schema of the original Inception module (left) and the SE-Inception module (right).}
\label{fig:seinception_module}
\end{figure}

The SE block can be integrated into standard architectures such as VGGNet \cite{simonyan_iclr2015vgg} by insertion after the non-linearity following each convolution. Moreover, the flexibility of the SE block means that it can be directly applied to transformations beyond standard convolutions. To illustrate this point, we develop SENets by incorporating SE blocks into several examples of more complex architectures, described next. 

\begin{figure}
\begin{center}
\includegraphics[height=65mm]{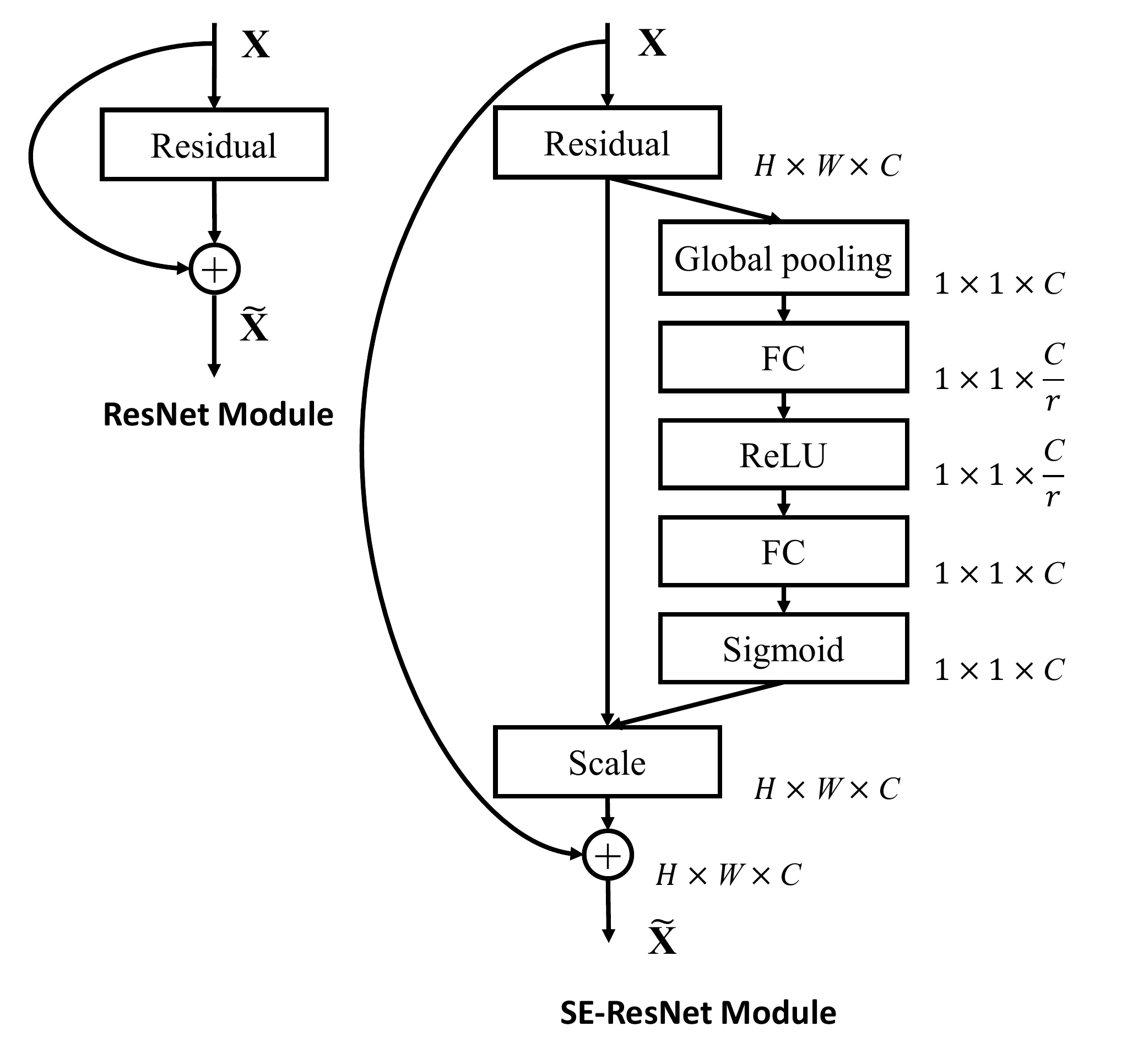}
\end{center}
\vspace{-2em}
\caption{The schema of the original Residual module (left) and the SE-ResNet module (right).}
\label{fig:seresnet_module}
\end{figure}

We first consider the construction of SE blocks for Inception networks \cite{szegedy_cvpr2015googlenet}.  Here, we simply take the transformation $\FF_{tr}$ to be an entire Inception module (see Fig.~\ref{fig:seinception_module}) and by making this change for each such module in the architecture, we obtain an \textit{SE-Inception} network. SE blocks can also be used directly with residual networks (Fig.~\ref{fig:seresnet_module} depicts the schema of an \textit{SE-ResNet} module). Here, the SE block transformation $\FF_{tr}$ is taken to be the non-identity branch of a residual module. \textit{Squeeze} and \textit{Excitation} both act before summation with the identity branch. Further variants that integrate SE blocks with ResNeXt \cite{xie_cvpr2017resnext}, Inception-ResNet \cite{szegedy_iclrw2016inceptionv4}, MobileNet \cite{howard_2017mobilenets} and ShuffleNet \cite{zhang_cvpr2018shufflenet} can be constructed by following similar schemes.  For concrete examples of SENet architectures, a detailed description of \mbox{SE-ResNet-50} and \mbox{SE-ResNeXt-50} is given in Table~\ref{model_structure}.  

One consequence of the flexible nature of the SE block is that there are several viable ways in which it could be integrated into these architectures.  Therefore, to assess sensitivity to the integration strategy used to incorporate SE blocks into a network architecture, we also provide ablation experiments exploring different designs for block inclusion in Section~\ref{sec:integration}.

%----------------------------------------------------------------------------
\section{Model and Computational Complexity}
\label{sec:modelcapacity}

For the proposed SE block design to be of practical use, it must offer a good trade-off between improved performance and increased model complexity. 
To illustrate the computational burden associated with the module, we consider a comparison between ResNet-50 and SE-ResNet-50 as an example.  ResNet-50 requires ${\sim}3.86$ GFLOPs in a single forward pass for a $224\times224$ pixel input image. Each SE block makes use of a global average pooling operation in the \textit{squeeze} phase and two small FC layers in the \textit{excitation} phase, followed by an inexpensive channel-wise scaling operation. In the aggregate, when setting the reduction ratio $r$ (introduced in Section~\ref{subsec:adaptive-recal}) to $16$, SE-ResNet-50 requires ${\sim}3.87$ GFLOPs, corresponding to a $0.26\%$ relative increase over the original ResNet-50. In exchange for this slight additional computational burden, the accuracy of \mbox{SE-ResNet-50} surpasses that of ResNet-50 and indeed, approaches that of a deeper ResNet-101 network requiring ${\sim}7.58$ GFLOPs (Table~\ref{tab:imagenet-results}). 

In practical terms, a single pass forwards and backwards through ResNet-50 takes $190$ ms, compared to $209$ ms for SE-ResNet-50 with a training minibatch of $256$ images (both timings are performed on a server with $8$ NVIDIA Titan X GPUs).  We suggest that this represents a reasonable runtime overhead, which may be further reduced as global pooling and small inner-product operations receive further optimisation in popular GPU libraries. Due to its importance for embedded device applications, we further benchmark CPU inference time for each model: for a $224\times 224$ pixel input image, ResNet-50 takes $164$ ms in comparison to $167$ ms for SE-ResNet-50. We believe that the small additional computational cost incurred by the SE block is justified by its contribution to model performance.

% -----------------------table ------------------------------------
%---------------------------------- table 5 -------------------------------
\begin{table*}[t]
\renewcommand\arraystretch{1.1}
\caption{(Left) ResNet-50 \cite{he_cvpr2016resnet}. (Middle) SE-ResNet-50.  (Right) SE-ResNeXt-50 with a 32$\times$4d template. The shapes and operations with specific parameter settings of a residual building block are listed inside the brackets and the number of stacked blocks in a stage is presented outside. The inner brackets following by \emph{fc} indicates the output dimension of the two fully connected layers in an SE module.}
\label{model_structure}
\vspace{-1.8em}
\begin{center}{\scalebox{0.96}{
\begin{tabular}{c|p{3.5cm}<{\centering}|p{4.9cm}<{\centering}|p{5.1cm}<{\centering}}
\hline
Output size & ResNet-50 & SE-ResNet-50 & SE-ResNeXt-50 ($32\times4$d) \\
\hline
$112\times112$ & \multicolumn{3}{c}{conv, $7\times7$, $64$, stride $2$} \\
\hline
\multirow{2}{*}{$56\times56$} & \multicolumn{3}{c}{max\;pool, $3\times3$, stride $2$} \\
\cline{2-4}
& 
  $\begin{bmatrix*}[l] {\rm conv}, 1\times 1, 64 \\ {\rm conv}, 3\times 3, 64 \\ {\rm conv}, 1\times 1, 256 \end{bmatrix*} \times 3$ &
  $\begin{bmatrix*}[l] {\rm conv}, 1\times 1, 64\\ {\rm conv}, 3\times 3, 64\\ {\rm conv}, 1\times 1, 256\\ fc, [16,256] \end{bmatrix*} \times 3$ &
  $\begin{bmatrix*}[l] {\rm conv}, 1\times 1, 128\\ {\rm conv}, 3\times 3, 128 & C=32\\ {\rm conv}, 1\times 1, 256\\ fc, [16,256] \end{bmatrix*} \times 3$ \\
\hline
$28\times28$ &
$\begin{bmatrix*}[l] {\rm conv}, 1\times 1, 128\\ {\rm conv}, 3\times 3, 128\\ {\rm conv}, 1\times 1, 512 \end{bmatrix*}\times 4$ &
$\begin{bmatrix*}[l] {\rm conv}, 1\times 1, 128\\ {\rm conv}, 3\times 3, 128\\ {\rm conv}, 1\times 1, 512\\ fc, [32, 512] \end{bmatrix*} \times 4$ & 
$\begin{bmatrix*}[l] {\rm conv}, 1\times 1, 256 &\\ {\rm conv}, 3\times 3, 256 & C=32\\ {\rm conv}, 1\times 1, 512\\ fc, [32,512] \end{bmatrix*} \times 4$ \\
\hline
$14\times14$ &
$\begin{bmatrix*}[l] {\rm conv}, 1\times 1, 256\\ {\rm conv}, 3\times 3, 256\\ {\rm conv}, 1\times 1, 1024 \end{bmatrix*} \times 6$ &
$\begin{bmatrix*}[l] {\rm conv}, 1\times 1, 256\\ {\rm conv}, 3\times 3, 256\\ {\rm conv}, 1\times 1, 1024\\ fc, [64, 1024] \end{bmatrix*} \times 6$ & 
$\begin{bmatrix*}[l] {\rm conv}, 1\times 1, 512 &\\ {\rm conv}, 3\times 3, 512 & C=32\\ {\rm conv}, 1\times 1, 1024\\ fc, [64, 1024] \end{bmatrix*} \times 6$ \\
\hline
7$\times$7 &
$\begin{bmatrix*}[l] {\rm conv}, 1\times 1, 512\\ {\rm conv}, 3\times 3, 512\\ {\rm conv}, 1\times1, 2048\end{bmatrix*} \times 3$ &
$\begin{bmatrix*}[l] {\rm conv}, 1\times 1, 512\\ {\rm conv}, 3\times 3, 512\\ {\rm conv}, 1\times1, 2048\\ fc, [128, 2048] \end{bmatrix*} \times 3$ & 
$\begin{bmatrix*}[l] {\rm conv}, 1\times 1, 1024 &\\ {\rm conv}, 3\times 3, 1024 & C=32\\ {\rm conv}, 1\times 1, 2048\\ fc, [128, 2048] \end{bmatrix*} \times 3$\\
\hline
$1\times1$ & \multicolumn{3}{c}{global average pool, $1000$-d $fc$, softmax} \\
\hline
\end{tabular}}}
\end{center}
\end{table*}
%---------------------------------- table -------------------------------
\begin{table*}[t]
\renewcommand\arraystretch{1.1}
\caption{\label{tab:imagenet-results}Single-crop error rates (\%) on the ImageNet validation set and complexity comparisons. The \textit{original} column refers to the results reported in the original papers (the results of ResNets are obtained from the website: https://github.com/Kaiminghe/deep-residual-networks). To enable a fair comparison, we re-train the baseline models and report the scores in the \textit{re-implementation} column. The \textit{SENet} column refers to the corresponding architectures in which SE blocks have been added. The numbers in brackets denote the performance improvement over the re-implemented baselines. $\dagger$ indicates that the model has been evaluated on the non-blacklisted subset of the validation set (this is discussed in more detail in \cite{szegedy_iclrw2016inceptionv4}), which may slightly improve results. VGG-16 and SE-VGG-16 are trained with batch normalization.}
\vspace{-1.8em}
\newcommand{\tabincell}[2]{\begin{tabular}{@{}#1@{}}#2\end{tabular}}
\begin{center} {\scalebox{0.97}{
\begin{tabular}{l|p{1.2cm}<{\centering}|p{1.2cm}<{\centering}|p{1.2cm}<{\centering}|p{1.2cm}<{\centering}|p{1.2cm}<{\centering}|p{1.6cm}<{\centering}|p{1.6cm}<{\centering}|p{1.2cm}<{\centering}}
\hline
\multirow{2}{*}{} & \multicolumn{2}{c|}{original} & \multicolumn{3}{c|}{re-implementation} & \multicolumn{3}{c}{SENet} \\
\cline{2-9}
& \tabincell{c}{top-1 err.} & \tabincell{c}{top-5 err.} & \tabincell{c}{top-1 err.} & \tabincell{c}{top-5 err.} & \tabincell{c}{GFLOPs} & \tabincell{c}{top-1 err.} & \tabincell{c}{top-5 err.} & \tabincell{c}{GFLOPs}\\
\hline
ResNet-50 \cite{he_cvpr2016resnet} & $24.7$ & $7.8$ & $24.80$ & $7.48$ & $3.86$ & $23.29_{(1.51)}$ & $6.62_{(0.86)}$  & $3.87$ \\
ResNet-101 \cite{he_cvpr2016resnet} & $23.6$ & $7.1$ & $23.17$ & $6.52$ & $7.58$ & $22.38_{(0.79)}$ & $6.07_{(0.45)}$ & $7.60$\\
ResNet-152 \cite{he_cvpr2016resnet} & $23.0$ & $6.7$ & $22.42$ & $6.34$ & $11.30$ & $21.57_{(0.85)}$ & $5.73_{(0.61)}$ & $11.32$\\
\hline
ResNeXt-50 \cite{xie_cvpr2017resnext} & $22.2$ & - & $22.11$ & $5.90$ & $4.24$ & $21.10_{(1.01)}$ & $5.49_{(0.41)}$ & $4.25$\\
ResNeXt-101 \cite{xie_cvpr2017resnext} & $21.2$ & $5.6$ & $21.18$ & $5.57$ & $7.99$ & $20.70_{(0.48)}$ & $5.01_{(0.56)}$ & $8.00$\\
\hline
VGG-16 \cite{simonyan_iclr2015vgg} & - & -  & $27.02$ & $8.81$ & $15.47$ & $25.22_{(1.80)}$  & $7.70_{(1.11)}$ & $15.48$ \\
BN-Inception \cite{ioffe_icml2015bn} & $25.2$ & $7.82$ & $25.38$ & $7.89$ & $2.03$ & $24.23_{(1.15)}$ & $7.14_{(0.75)}$ & $2.04$\\
Inception-ResNet-v2 \cite{szegedy_iclrw2016inceptionv4} & $19.9^\dagger$ & $4.9^\dagger$ & $20.37$ & $5.21$ & $11.75$ & $19.80_{(0.57)}$ & $4.79_{(0.42)}$  & $11.76$\\
\hline
\end{tabular}}}
\end{center}
\end{table*}
%------------------------------------------------------------------

We next consider the additional parameters introduced by the proposed SE block.  These additional parameters result solely from the two FC layers of the gating mechanism and therefore constitute a small fraction of the total network capacity.  Concretely, the total number introduced by the weight parameters of these FC layers is given by: 
\vspace{-1mm}
\begin{equation}
\label{model_capacity}
\frac{2}{r} \sum_{s=1}^S N_s \cdot {C_s}^2,
\end{equation}
where $r$ denotes the reduction ratio, $S$ refers to the number of stages (a stage refers to the collection of blocks operating on feature maps of a common spatial dimension), $C_s$ denotes the dimension of the output channels and $N_s$ denotes the number of repeated blocks for stage $s$ (when bias terms are used in FC layers, the introduced parameters and computational cost are typically negligible).
\mbox{SE-ResNet-50} introduces \mbox{${\sim}2.5$ million} additional parameters beyond the \mbox{${\sim}25$ million} parameters required by \mbox{ResNet-50}, corresponding to a ${\sim}10\%$ increase. In practice, the majority of these parameters come from the final stage of the network, where the excitation operation is performed across the greatest number of channels. However, we found that this comparatively costly final stage of SE blocks could be removed at only a small cost in performance (${<}0.1\%$ top-$5$ error on ImageNet) reducing the relative parameter increase to ${\sim}4\%$, which may prove useful in cases where parameter usage is a key consideration (see Section~\ref{subsec:ablation-stages} and~\ref{subsection:role-of-excitation} for further discussion).
\section{Experiments}\label{sec:exp-results}

In this section, we conduct experiments to investigate the effectiveness of SE blocks across a range of tasks, datasets and model architectures.

%------------------table---------------------------------------------
%------------------------------------- table --------------------------------------------
\begin{table*}[t]
\renewcommand\arraystretch{1.1} 
\caption{Single-crop error rates (\%) on the ImageNet validation set and complexity comparisons. MobileNet refers to ``1.0 MobileNet-224" in \cite{howard_2017mobilenets} and ShuffleNet refers to ``ShuffleNet $1\times (g = 3)$'' 
in \cite{zhang_cvpr2018shufflenet}. The numbers in brackets denote the performance improvement over the re-implementation.}
\label{tab:smallnet-results}
\vspace{-1.8em}
\newcommand{\tabincell}[2]{\begin{tabular}{@{}#1@{}}#2\end{tabular}}
\begin{center}{\scalebox{0.92}{
  \begin{tabular}{l|p{1.2cm}<{\centering}|p{1.2cm}<{\centering}|p{1.2cm}<{\centering}|p{1.2cm}<{\centering}|p{1.2cm}<{\centering}|p{1.3cm}<{\centering}|p{1.3cm}<{\centering}|p{1.3cm}<{\centering}|p{1.2cm}<{\centering}|p{1.3cm}<{\centering}}
  \hline
  \multirow{2}{*}{} & \multicolumn{2}{c|}{original} & \multicolumn{4}{c|}{re-implementation} & \multicolumn{4}{c}{SENet} \\
  \cline{2-11}
  & top-1 err. & \tabincell{c}{top-5 err.} & \tabincell{c}{top-1 err.} & \tabincell{c}{top-5 err.} & MFLOPs & \tabincell{c}{Params} & \tabincell{c}{top-1 err.} & \tabincell{c}{top-5 err.} & MFLOPs & \tabincell{c}{Params} \\
  \hline
  MobileNet \cite{howard_2017mobilenets} & $29.4$ & - & $28.4$ & $9.4$ & $569$ & $4.2$M & $25.3_{(3.1)}$ & $7.7_{(1.7)}$  & $572$ & $4.7$M \\
  ShuffleNet \cite{zhang_cvpr2018shufflenet} & $32.6$ & - & $32.6$ & $12.5$ & $140$ & $1.8$M & $31.0_{(1.6)}$ & $11.1_{(1.4)}$ & $142$ & $2.4$M \\ 
  \hline
    \end{tabular}}}
  \end{center}
\end{table*}
%-------------------------------------------------------------------
\begin{figure*}
\begin{center}
\includegraphics[width=\textwidth]{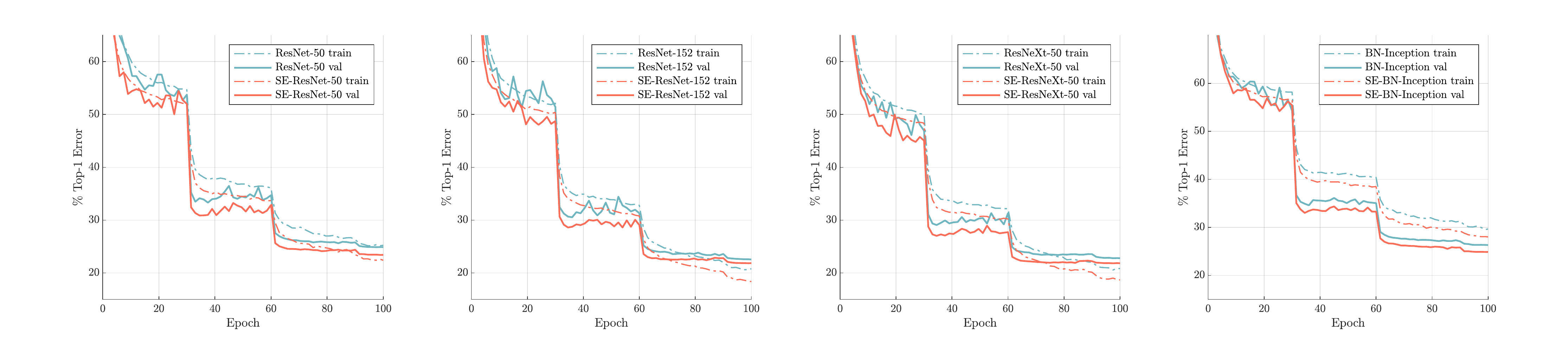}
\end{center}
\vspace{-1.5em}
\caption[Training curves]{Training baseline architectures and their SENet counterparts on ImageNet. SENets exhibit improved optimisation characteristics and produce consistent gains in performance which are sustained throughout the training process. 
}
\label{fig:curves}
\end{figure*}

\subsection{Image Classification}\label{subsec:imagenet}

To evaluate the influence of SE blocks, we first perform experiments on the ImageNet $2012$ dataset \cite{russakovsky_ijcv2015imagenet} which comprises $1.28$ million training images and $50$K validation images from $1000$ different classes. We train networks on the training set and report the top-$1$ and top-$5$ error on the validation set.

Each baseline network architecture and its corresponding SE counterpart are trained with identical optimisation schemes.  We follow standard practices and perform data augmentation with random cropping using scale and aspect ratio \cite{szegedy_cvpr2015googlenet} to a size of $224\times 224$ pixels (or $299\times 299$ for Inception-ResNet-v2 \cite{szegedy_iclrw2016inceptionv4} and SE-Inception-ResNet-v2) and perform random horizontal flipping. Each input image is normalised through mean RGB-channel subtraction. All models are trained on our distributed learning system \textit{ROCS} which is designed to handle efficient parallel training of large networks. Optimisation is performed using synchronous SGD with momentum $0.9$ and a minibatch size of $1024$. The initial learning rate is set to $0.6$ and decreased by a factor of $10$ every $30$ epochs. Models are trained for $100$ epochs from scratch, using the weight initialisation strategy described in \cite{he_iccv2015prelu}. The reduction ratio $r$ (in Section~\ref{subsec:adaptive-recal}) is set to $16$ by default (except where stated otherwise). 

When evaluating the models we apply centre-cropping so that $224\times 224$ pixels are cropped from each image, after its shorter edge is first resized to $256$ ($299\times 299$ from each image whose shorter edge is first resized to $352$ for Inception-ResNet-v2 and SE-Inception-ResNet-v2).

\vspace{5pt}
\noindent\textbf{Network depth.}
We begin by comparing SE-ResNet against ResNet architectures with different depths and report the results in Table~\ref{tab:imagenet-results}. We observe that SE blocks consistently improve performance across different depths with an extremely small increase in computational complexity.  Remarkably, SE-ResNet-50 achieves a single-crop \mbox{top-5} validation error of $6.62\%$, exceeding ResNet-50 (7.48\%) by 0.86\% and approaching the performance achieved by the much deeper ResNet-101 network (6.52\% top-5 error) with only half of the total computational burden ($3.87$ GFLOPs vs. $7.58$ GFLOPs). This pattern is repeated at greater depth, where SE-ResNet-101 ($6.07\%$ top-$5$ error) not only matches, but outperforms the deeper ResNet-152 network ($6.34\%$ top-5 error) by $0.27\%$. While it should be noted that the SE blocks themselves add depth, they do so in an extremely computationally efficient manner and yield good returns even at the point at which extending the depth of the base architecture achieves diminishing returns. Moreover, we see that the gains are consistent across a range of different network depths, suggesting that the improvements induced by SE blocks may be complementary to those obtained by simply increasing the depth of the base architecture.

\vspace{5pt}
\noindent\textbf{Integration with modern architectures.}
We next study the effect of integrating SE blocks with two further state-of-the-art architectures, Inception-ResNet-v2 \cite{szegedy_iclrw2016inceptionv4} and ResNeXt (using the setting of $32\times 4$d) \cite{xie_cvpr2017resnext}, both of which introduce additional computational building blocks into the base network.  We construct SENet equivalents of these networks, SE-Inception-ResNet-v2 and SE-ResNeXt (the configuration of SE-ResNeXt-50 is given in Table~\ref{model_structure}) and report results in  Table~\ref{tab:imagenet-results}.  As with the previous experiments, we observe significant performance improvements induced by the introduction of SE blocks into both architectures. In particular, \mbox{SE-ResNeXt-50} has a top-5 error of $5.49$\% which is superior to both its direct counterpart ResNeXt-50 ($5.90\%$ top-5 error) as well as the deeper ResNeXt-101 ($5.57\%$ top-5 error), a model which has almost twice the total number of parameters and computational overhead. We note a slight difference in performance between our re-implementation of {\mbox Inception-ResNet-v2} and the result reported in \cite{szegedy_iclrw2016inceptionv4}. However, we observe a similar trend with regard to the effect of SE blocks, finding that SE counterpart ($4.79\%$ \mbox{top-5} error) outperforms our reimplemented \mbox{Inception-ResNet-v2} baseline ($5.21\%$ top-5 error) by $0.42\%$ as well as the reported result in \cite{szegedy_iclrw2016inceptionv4}.  

We also assess the effect of SE blocks when operating on \textit{non-residual} networks by conducting experiments with the VGG-16 \cite{simonyan_iclr2015vgg} and BN-Inception architecture \cite{ioffe_icml2015bn}.  To facilitate the training of VGG-16 from scratch, we add Batch Normalization layers after each convolution. We use identical training schemes for both VGG-16 and SE-VGG-16. The results of the comparison are shown in Table \ref{tab:imagenet-results}. Similarly to the results reported for the residual baseline architectures, we observe that SE blocks bring improvements in performance on the non-residual settings.

To provide some insight into influence of SE blocks on the optimisation of these models, example training curves for runs of the baseline architectures 
and their respective SE counterparts are depicted in Fig.~\ref{fig:curves}.
We observe that SE blocks yield a steady improvement throughout the optimisation procedure.  Moreover, this trend is fairly consistent across a range of network architectures considered as baselines.

\vspace{5pt}
\noindent\textbf{Mobile setting.} Finally, we consider two representative architectures from the class of mobile-optimised networks, MobileNet~\cite{howard_2017mobilenets} and ShuffleNet~\cite{zhang_cvpr2018shufflenet}. For these experiments, we used a minibatch size of 256 and slightly less aggressive data augmentation and regularisation as in~\cite{zhang_cvpr2018shufflenet}. We trained the models across 8 GPUs using SGD with momentum (set to 0.9) and an initial learning rate of 0.1 which was reduced by a factor of 10 each time the validation loss plateaued. The total training process required \mbox{$\sim400$} epochs (enabling us to reproduce the baseline performance of~\cite{zhang_cvpr2018shufflenet}). The results reported in Table~\ref{tab:smallnet-results} show that SE blocks consistently improve the accuracy by a large margin at a minimal increase in computational cost.

\vspace{5pt}
\noindent\textbf{Additional datasets.} We next investigate whether the benefits of SE blocks generalise to datasets beyond ImageNet. We perform experiments with several popular baseline architectures and techniques (\mbox{ResNet-110} \cite{he_eccv2016preact}, \mbox{ResNet-164} \cite{he_eccv2016preact}, \mbox{WideResNet-16-8} \cite{zagoruyko_bmvc2016wrn}, \mbox{Shake-Shake} \cite{xavier_2017shake} and \mbox{Cutout} \cite{terrance_2017cutout}) on the CIFAR-$10$ and CIFAR-$100$ datasets \cite{krizhevsky2009learning}.  These comprise a collection of 50k training and 10k test $32 \times 32$ pixel RGB images, labelled with 10 and 100 classes respectively.  The integration of SE blocks into these networks follows the same approach that was described in Section~\ref{se_inception_resnet}.  Each baseline and its SENet counterpart are trained with standard data augmentation strategies~\cite{lin_iclr2014nin,huang_eccv2016sdepth}. During training, images are randomly horizontally flipped and zero-padded on each side with four pixels before taking a  random $32 \times 32$ crop. Mean and standard deviation normalisation is also applied. The setting of the training hyperparameters (e.g. minibatch size, initial learning rate, weight decay) match those suggested by the original papers. We report the performance  of each baseline and its SENet counterpart on CIFAR-$10$ in Table~\ref{tab:cifar10} and performance on CIFAR-$100$ in Table~\ref{tab:cifar100}.  We observe that in every comparison SENets outperform the baseline architectures, suggesting that the benefits of SE blocks are not confined to the ImageNet dataset.

%--------------------table------------------------------------------
\begin{table}
\renewcommand\arraystretch{1.1}
\small
\caption{Classification error (\%) on CIFAR-10. }
\label{tab:cifar10}
\vspace{-1.8em}
\begin{center}{\scalebox{0.91}{
\begin{tabular}{l|p{1.1cm}<{\centering}|p{1.1cm}<{\centering}}
\hline
& original  & SENet  \\
\hline
ResNet-110 \cite{he_eccv2016preact} & $6.37$ & $5.21$ \\
ResNet-164 \cite{he_eccv2016preact} & $5.46$ & $4.39$ \\
WRN-16-8 \cite{zagoruyko_bmvc2016wrn} & $4.27$ & $3.88$ \\
Shake-Shake 26 2x96d \cite{xavier_2017shake} + Cutout \cite{terrance_2017cutout} & $2.56$ & $2.12$\\
\hline
\end{tabular}}}
\end{center}
\end{table}
\begin{table}
\renewcommand\arraystretch{1.1}
\small
\caption{Classification error (\%) on CIFAR-100. }
\label{tab:cifar100}
\vspace{-1.8em}
\begin{center}{\scalebox{0.91}{
\begin{tabular}{l|p{1.1cm}<{\centering}|p{1.1cm}<{\centering}}
\hline
& original  & SENet  \\
\hline
ResNet-110 \cite{he_eccv2016preact} & $26.88$ & $23.85$ \\
ResNet-164 \cite{he_eccv2016preact} & $24.33$ & $21.31$ \\
WRN-16-8 \cite{zagoruyko_bmvc2016wrn} & $20.43$ & $19.14$ \\
Shake-Even 29 2x4x64d \cite{xavier_2017shake} + Cutout \cite{terrance_2017cutout} & $15.85$ & $15.41$ \\
\hline
\end{tabular}}}
\end{center}
\end{table}

%------------------------------------- table --------------------------------------------
\begin{table}[t]
\renewcommand\arraystretch{1.1}
\small
\caption{Single-crop error rates (\%) on Places365 validation set.}
\label{tab:places-results}
\vspace{-1.5em}
\begin{center}{\scalebox{0.95}{
\begin{tabular}{l|p{1.3cm}<{\centering}|p{1.3cm}<{\centering}}
\hline
& top-1 err. & top-5 err.\\
\hline
Places-365-CNN \cite{shen_places365} & $41.07$ & $11.48$ \\
ResNet-152 (ours) & $41.15$ & $11.61$ \\
SE-ResNet-152 & \textbf{40.37} & \textbf{11.01} \\
\hline
\end{tabular}}}
\end{center}
\end{table}

%-------------------------------------------------------------------

\subsection{Scene Classification}

We also conduct experiments on the Places365-Challenge dataset \cite{zhou_pami2017places} for scene classification. This dataset comprises $8$ million training images and $36,500$ validation images across $365$ categories. Relative to classification, the task of scene understanding offers an alternative assessment of a model's ability to generalise well and handle abstraction.  This is because it often requires the model to handle more complex data associations and to be robust to a greater level of appearance variation.

We opted to use ResNet-152 as a strong baseline to assess the effectiveness of SE blocks and follow the training and evaluation protocols described in \cite{shen_places365, shen_eccv2016}. In these experiments, models are trained from scratch. We report the results in Table~\ref{tab:places-results}, comparing also with prior work. We observe that SE-ResNet-152 ($11.01\%$ {\mbox top-5} error) achieves a lower validation error than ResNet-152 ($11.61\%$ top-5 error), providing evidence that SE blocks can also yield improvements for scene classification. This SENet surpasses the previous state-of-the-art model Places-365-CNN \cite{shen_places365} which has a top-5 error of $11.48\%$ on this task.

%------------------------------------- table --------------------------------------------
\begin{table}[t]
\renewcommand{\arraystretch}{1.1}
\caption{Faster R-CNN object detection results (\%) on COCO {\it minival} set.}
\label{tab:coco-results}
\vspace{-1.5em}
\small
\begin{center}{\scalebox{0.96}{
\begin{tabular}{l|p{2.0cm}<{\centering} |p{2.3cm}<{\centering}}
\hline
 & AP@IoU=$0.5$ & AP \\
\hline
ResNet-50 & 57.9 & 38.0\\
SE-ResNet-50 & 61.0 & 40.4 \\
\hline
ResNet-101 & 60.1 & 39.9 \\
SE-ResNet-101 & 62.7 & 41.9 \\
\hline
\end{tabular}}}
\end{center}
\end{table}	

\subsection{Object Detection on COCO}
\label{coco_det}

We further assess the generalisation of SE blocks on the task of object detection using the COCO dataset \cite{lin_eccv2014coco}. As in previous work \cite{xie_cvpr2017resnext}, we use the {\it minival} protocol, i.e., training the models on the union of the $80$k training set and a $35$k val subset and evaluating on the remaining $5$k val subset.  Weights are initialised by the parameters of the model trained on the ImageNet dataset. We use the Faster \mbox{R-CNN} \cite{ren_nips2015fasterrcnn} detection framework as the basis for evaluating our models and follow the hyperparameter setting described in \cite{Detectron2018} (i.e., end-to-end training with the '2x' learning schedule). Our goal is to evaluate the effect of replacing the trunk architecture (ResNet) in the object detector with \mbox{SE-ResNet}, so that any changes in performance can be attributed to better representations.
Table~\ref{tab:coco-results} reports the validation set performance of the object detector using \mbox{ResNet-50}, \mbox{ResNet-101} and their SE counterparts as trunk architectures. \mbox{SE-ResNet-50} outperforms \mbox {ResNet-50} by $2.4\%$ (a relative $6.3\%$ improvement) on COCO's standard AP metric and by $3.1\%$ on AP@IoU=$0.5$. SE blocks also benefit the deeper \mbox{ResNet-101} architecture achieving a $2.0\%$ improvement ($5.0\%$ relative improvement) on the AP metric.
In summary, this set of experiments demonstrate the generalisability of SE blocks. The induced improvements can be realised across a broad range of architectures, tasks and datasets.

\vspace{5pt}
\subsection{ILSVRC 2017 Classification Competition} \label{sec:ilsvrc}
%---------------------------------- table-------------------------------
\begin{table}[t!]
\renewcommand\arraystretch{1.1}
\caption{Single-crop error rates (\%) of state-of-the-art CNNs on ImageNet validation set with crop sizes $224\times224$ and $320\times320$ / $299\times299$.}
\label{tab:challenge-results}
\vspace{-1.8em}
\begin{center}
\begin{tabular}{l|p{0.8cm}<{\centering}|p{0.8cm}<{\centering}|p{0.8cm}<{\centering}|p{0.8cm}<{\centering}}
\hline
\multirow{2}{*}{} & \multicolumn{2}{c|}{$224 \times 224$} & \multicolumn{2}{p{1.8cm}<{\centering}}{$320 \times 320$ / $299 \times 299$} \\
\cline{2-5}
& top-1 err. & top-5 err. & top-1 err. & top-5 err. \\
\hline
ResNet-152 \cite{he_cvpr2016resnet} & $23.0$ & $6.7$ & $21.3$ & $5.5$ \\
ResNet-200 \cite{he_eccv2016preact} & $21.7$ & $5.8$ & $20.1$ & $4.8$ \\
Inception-v3 \cite{szegedy_cvpr2016inceptionv3} & - & - & $21.2$ & $5.6$ \\
Inception-v4 \cite{szegedy_iclrw2016inceptionv4} & - & - & $20.0$ & $5.0$ \\
Inception-ResNet-v2 \cite{szegedy_iclrw2016inceptionv4} & - & - & $19.9$ & $4.9$ \\
ResNeXt-101 (64 $\times$ 4d) \cite{xie_cvpr2017resnext}  & $20.4$ & $5.3$ & $19.1$ & $4.4$ \\
DenseNet-264 \cite{huang_cvpr2017dns} & $22.15$ & $6.12$ & - & - \\
Attention-92 \cite{wangfei_cvpr2017resattent} & - & - & $19.5$ & $4.8$ \\
PyramidNet-200 \cite{han_cvpr2017dprn} & $20.1$ & $5.4$ & $19.2$ & $4.7$ \\
DPN-131 \cite{chen_nips2017dpn} & $19.93$ & $5.12$ & $18.55$ & $4.16$ \\
\textbf{SENet-154} & \textbf{18.68} & \textbf{4.47} & $\mathbf{17.28}$ & $\mathbf{3.79}$ \\
\hline
\end{tabular}
\end{center}
\end{table}

%---------------------------------- table-------------------------------
\begin{table}[t!]
\renewcommand\arraystretch{1.1}
\caption{Comparison (\%) with state-of-the-art CNNs on ImageNet validation set using larger crop sizes/additional training data.  $^\dagger$This model was trained with a crop size of $320 \times 320$.}
\label{tab:modern-results}
\vspace{-1.5em}
\begin{center}
\begin{tabular}{l|p{0.6cm}<{\centering}|p{0.75cm}<{\centering}|p{0.75cm}<{\centering}|p{0.75cm}<{\centering}}
\hline
 & extra data & crop size & top-1 err. & top-5 err. \\
\hline
Very Deep PolyNet \cite{zhang_cvpr2017polynet} & - & $331$ & $18.71$ & $4.25$ \\
NASNet-A (6 @ 4032) \cite{zoph_cvpr2018}  & - & $331$ & $17.3$ & $3.8$ \\
PNASNet-5 (N=4,F=216) \cite{liu_eccv2018progressive} & - & $331$ & $17.1$ & $3.8$ \\
SENet-154$^\dagger$ & - & $320$ & $16.88$ & $3.58$ \\
AmoebaNet-C \cite{cubuk_2018autoaugment} & - & $331$ & $16.5$ & $3.5$ \\
ResNeXt-101 $32 \times 48$d \cite{mahajan_eccv2018exploring} & \checkmark & $224$ & $14.6$ & $2.4$ \\
\hline
\end{tabular}
\end{center}
\end{table}

SENets formed the foundation of our submission to the ILSVRC competition where we achieved first place. Our winning entry comprised a small ensemble of SENets that employed a standard multi-scale and multi-crop fusion strategy to obtain a top-5 error of $2.251\%$ on the test set.  As part of this submission, we constructed an additional model, {\mbox {\it SENet-154}}, by integrating SE blocks with a modified ResNeXt \cite{xie_cvpr2017resnext} (the details of the architecture are provided in Appendix).  We compare this model with prior work on the ImageNet validation set in Table \ref{tab:challenge-results} using standard crop sizes ($224 \times 224$ and $320 \times 320$).  We observe that SENet-$154$ achieves a top-1 error of $18.68\%$  and a top-5 error of $4.47\%$ using a $224 \times 224$ centre crop evaluation, which represents the strongest reported result.

Following the challenge there has been a great deal of further progress on the ImageNet benchmark. For comparison, we include the strongest results that we are currently aware of %among the both published and unpublished literature 
in Table~\ref{tab:modern-results}.  The best performance using only ImageNet data was recently reported by \cite{cubuk_2018autoaugment}.  This method uses reinforcement learning to develop new policies for data augmentation during training to improve the performance of the architecture searched by \cite{real_2018a}.  The best overall performance was reported by \cite{mahajan_eccv2018exploring} using a ResNeXt-$101$ $32 \times 48d$ architecture. This was achieved by pretraining their model on approximately one billion weakly labelled images and finetuning on ImageNet.  The improvements yielded by more sophisticated data augmentation \cite{cubuk_2018autoaugment} and extensive pretraining \cite{mahajan_eccv2018exploring} may be complementary to our proposed changes to the network architecture.
\section{Ablation Study}\label{ablation_study}

In this section we conduct ablation experiments to gain a better understanding of the effect of using different configurations on components of the SE blocks.  All ablation experiments are performed on the ImageNet dataset on a single machine (with 8 GPUs). \mbox{ResNet-50} is used as the backbone architecture. We found empirically that on ResNet architectures, removing the biases of the FC layers in the excitation operation facilitates the modelling of channel dependencies, and use this configuration in the following experiments.
The data augmentation strategy follows the approach described in Section~\ref{subsec:imagenet}.  
To allow us to study the upper limit of performance for each variant, the learning rate is initialised to 0.1 and training continues until the validation loss plateaus\footnote{For reference, training with a 270 epoch fixed schedule (reducing the learning rate at 125, 200 and 250 epochs) achieves \mbox{top-1} and \mbox{top-5} error rates for \mbox{ResNet-50} and \mbox{SE-ResNet-50} of ($23.21\%$, $6.53\%$) and ($22.20\%$, $6.00\%$) respectively.} (${\sim}300$ epochs in total). The learning rate is then reduced by a factor of 10 and then this process is repeated (three times in total).  Label-smoothing regularisation  \cite{szegedy_cvpr2016inceptionv3} is used during training.

%------------------------------------- table--------------------------------------------
\begin{table}[t]
\renewcommand\arraystretch{1.1}
\small
\caption{Single-crop error rates (\%) on ImageNet and parameter sizes for \mbox{SE-ResNet-50} at different reduction ratios. Here, \textit{original} refers to \mbox{ResNet-50}.}
\label{table_reduction_ratio}
\vspace{-1.5em}
\begin{center}{\scalebox{0.95}{
\begin{tabular}{c|p{1.3cm}<{\centering}|p{1.3cm}<{\centering}|p{1.5cm}<{\centering}}
\hline
Ratio $r$ & top-1 err. & top-5 err. & Params\\
\hline
$2$ & $22.29$ & $6.00$ & $45.7$M\\
$4$ & $22.25$ & $6.09$ & $35.7$M\\
$8$ & $22.26$ & $5.99$ & $30.7$M\\
$16$ & $22.28$ & $6.03$ & $28.1$M\\
$32$ & $22.72$ & $6.20$ & $26.9$M\\
\hline
original & $23.30$ & $6.55$ & $25.6$M \\
\hline
\end{tabular}}}
\end{center}
\end{table}

\subsection{Reduction ratio}\label{subsec:reduction} The reduction ratio $r$ introduced in Eqn.~\ref{model_capacity} is a hyperparameter which allows us to vary the capacity and computational cost of the SE blocks in the network. To investigate the trade-off between performance and computational cost mediated by this hyperparameter, we conduct experiments with \mbox{SE-ResNet-50} for a range of different $r$ values. The comparison in Table \ref{table_reduction_ratio} shows that performance is robust to a range of reduction ratios. Increased complexity does not improve performance monotonically while a smaller ratio dramatically increases the parameter size of the model.  
Setting $r = 16$ achieves a good balance between accuracy and complexity. In practice, using an identical ratio throughout a network may not be optimal (due to the distinct roles performed by different layers), so further improvements may be achievable by tuning the ratios to meet the needs of a given base architecture.

\subsection{Squeeze Operator} \label{subsec:ablation-squeeze}

We examine the significance of using global average pooling as opposed to global max pooling as our choice of squeeze operator (since this worked well, we did not consider more sophisticated alternatives).  The results are reported in Table~\ref{tab:squeeze}.  While both max and average pooling are effective, average pooling achieves slightly better performance, justifying its selection as the basis of the squeeze operation.  However, we note that the performance of SE blocks is fairly robust to the choice of specific aggregation operator.

\subsection{Excitation Operator} \label{subsec:ablation-excitation}

We next assess the choice of non-linearity for the excitation mechanism.  We consider two further options: ReLU and tanh, and experiment with replacing the sigmoid with these alternative non-linearities. The results are reported in Table~\ref{tab:non-linearities}. We see that exchanging the sigmoid for tanh slightly worsens performance, while using ReLU is dramatically worse and in fact causes the performance of \mbox{SE-ResNet-50} to drop below that of the \mbox{ResNet-50} baseline.  This suggests that for the SE block to be effective, careful construction of the excitation operator is important.
%---------------table---------------------------------------------
%------------------------------------- table --------------------------------------------
\begin{table}[t]
\renewcommand\arraystretch{1.1}
\small
\caption{Effect of using different squeeze operators in SE-ResNet-50 on ImageNet (error rates \%).}
\label{tab:squeeze}
\vspace{-1.5em}
\begin{center}{\scalebox{0.95}{
\begin{tabular}{p{2.5cm}<{\centering}|p{1.3cm}<{\centering}|p{1.3cm}<{\centering}}
\hline
Squeeze & top-1 err. & top-5 err.\\
\hline
Max & 22.57 & 6.09 \\
Avg & $\mathbf{22.28}$ & $\mathbf{6.03}$ \\
\hline
\end{tabular}}}
\end{center}
\end{table}

%------------------------------------- table --------------------------------------------
\begin{table}[t]
\renewcommand\arraystretch{1.1}
\small
\caption{Effect of using different non-linearities for the excitation operator in SE-ResNet-50 on ImageNet (error rates \%).}
\label{tab:non-linearities}
\vspace{-1.5em}
\begin{center}{\scalebox{0.95}{
\begin{tabular}{p{2.5cm}<{\centering}|p{1.3cm}<{\centering}|p{1.3cm}<{\centering}}
\hline
Excitation & top-1 err. & top-5 err.\\
\hline
ReLU & 23.47 & 6.98 \\
Tanh & 23.00  &  6.38  \\
Sigmoid & $\mathbf{22.28}$ & $\mathbf{6.03}$ \\
\hline
\end{tabular}}}
\end{center}
\end{table}

%------------------------------------- table --------------------------------------------
\begin{table}[b]
\renewcommand\arraystretch{1.1}
\small
\caption{Effect of integrating SE blocks with ResNet-50 at different stages on ImageNet (error rates \%).}
\label{tab:stages}
\vspace{-1.5em}
\begin{center}{\scalebox{0.95}{
\begin{tabular}{l|p{1.4cm}<{\centering}|p{1.4cm}<{\centering}|p{1.2cm}<{\centering}|p{1.2cm}<{\centering}}
\hline
Stage & top-1 err. & top-5 err. & GFLOPs & Params\\
\hline
ResNet-50 & 23.30 & 6.55  & 3.86 & 25.6M\\
SE\_Stage\_2 & 23.03 & 6.48 & 3.86 & 25.6M\\
SE\_Stage\_3 & 23.04 &  6.32  & 3.86 & 25.7M\\
SE\_Stage\_4 & 22.68 & 6.22 & 3.86 & 26.4M\\
\hline
SE\_All          & 22.28 & 6.03 & 3.87 & 28.1M\\
\hline
\end{tabular}}}
\end{center}
\end{table}
%-----------------------------------------------------------------
\subsection{Different stages} \label{subsec:ablation-stages}

We explore the influence of SE blocks at different stages by integrating SE blocks into ResNet-50, one stage at a time. Specifically, we add SE blocks to the intermediate stages: stage\_2, stage\_3 and stage\_4, and report the results in Table~\ref{tab:stages}.  We observe that SE blocks bring performance benefits when introduced at each of these stages of the architecture. Moreover, the gains induced by SE blocks at different stages are complementary, in the sense that they can be combined effectively to further bolster network performance.

\subsection{Integration strategy} \label{sec:integration}

\begin{figure*}
\begin{center}
	\subfigure[Residual block]{\includegraphics[width=.16 \textwidth]{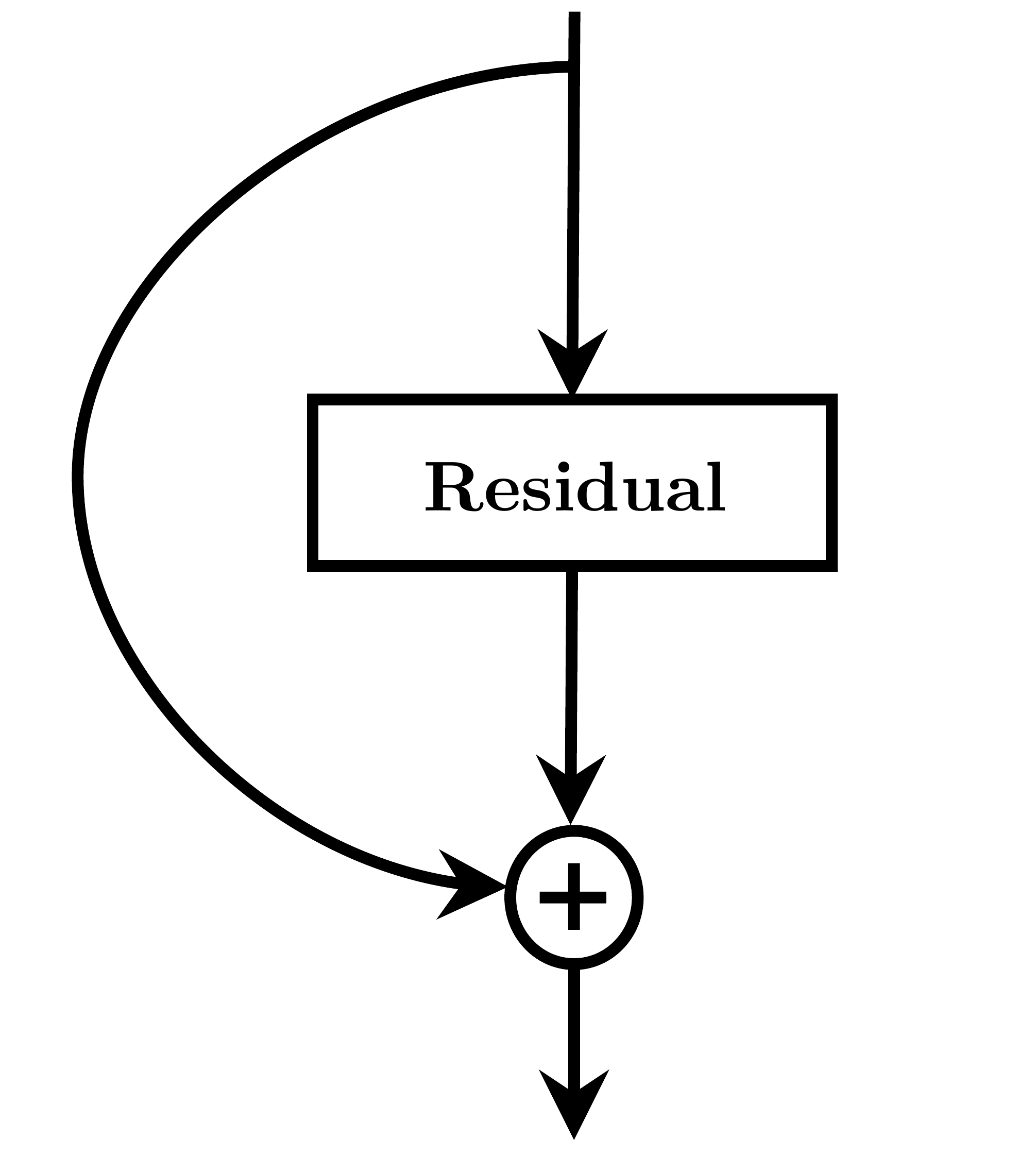}} \hspace{3mm}
	\subfigure[Standard SE block]{\includegraphics[width=.16 \textwidth]{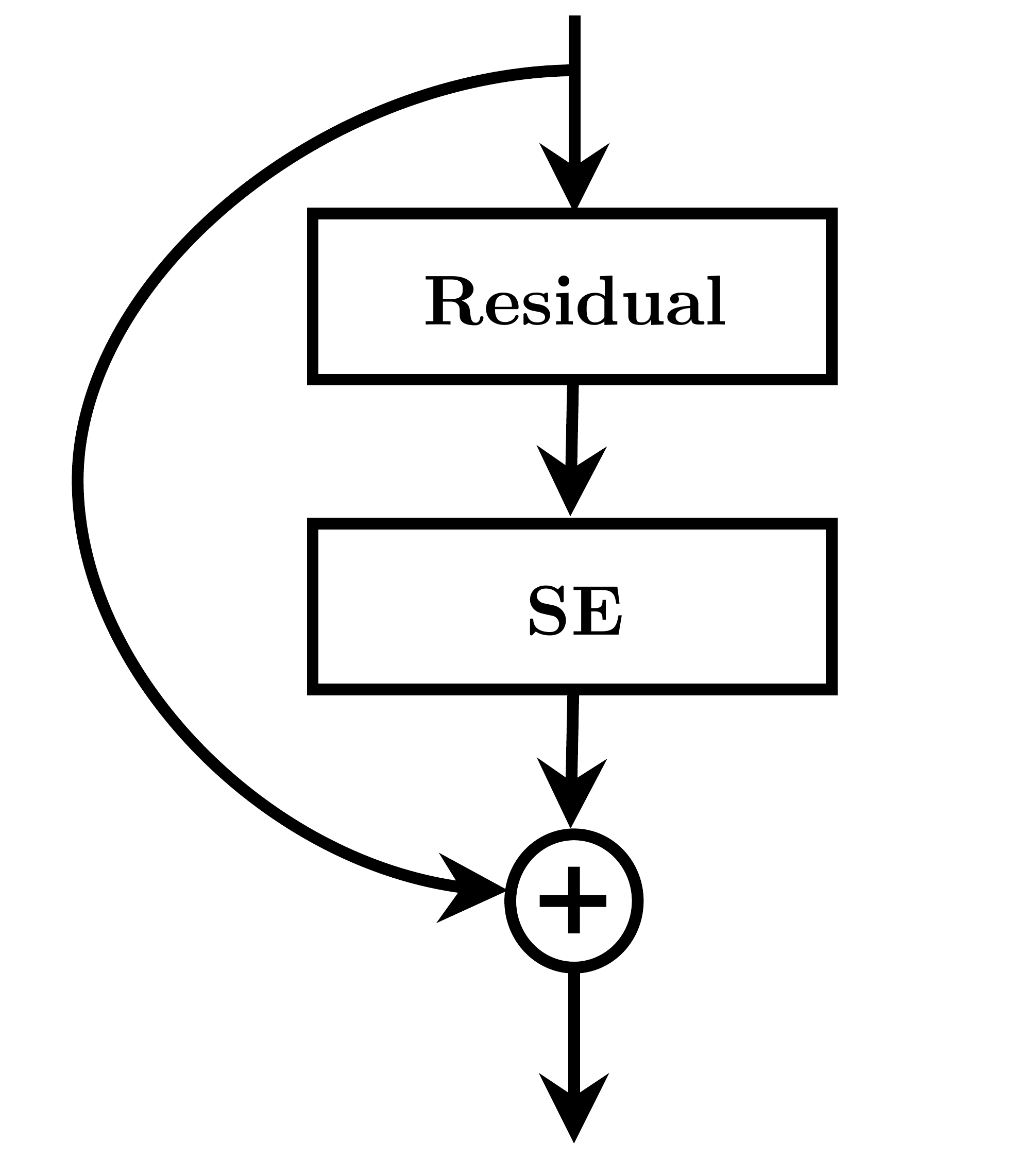}} \hspace{3mm}
	\subfigure[SE-PRE block]{\includegraphics[width=.16 \textwidth]{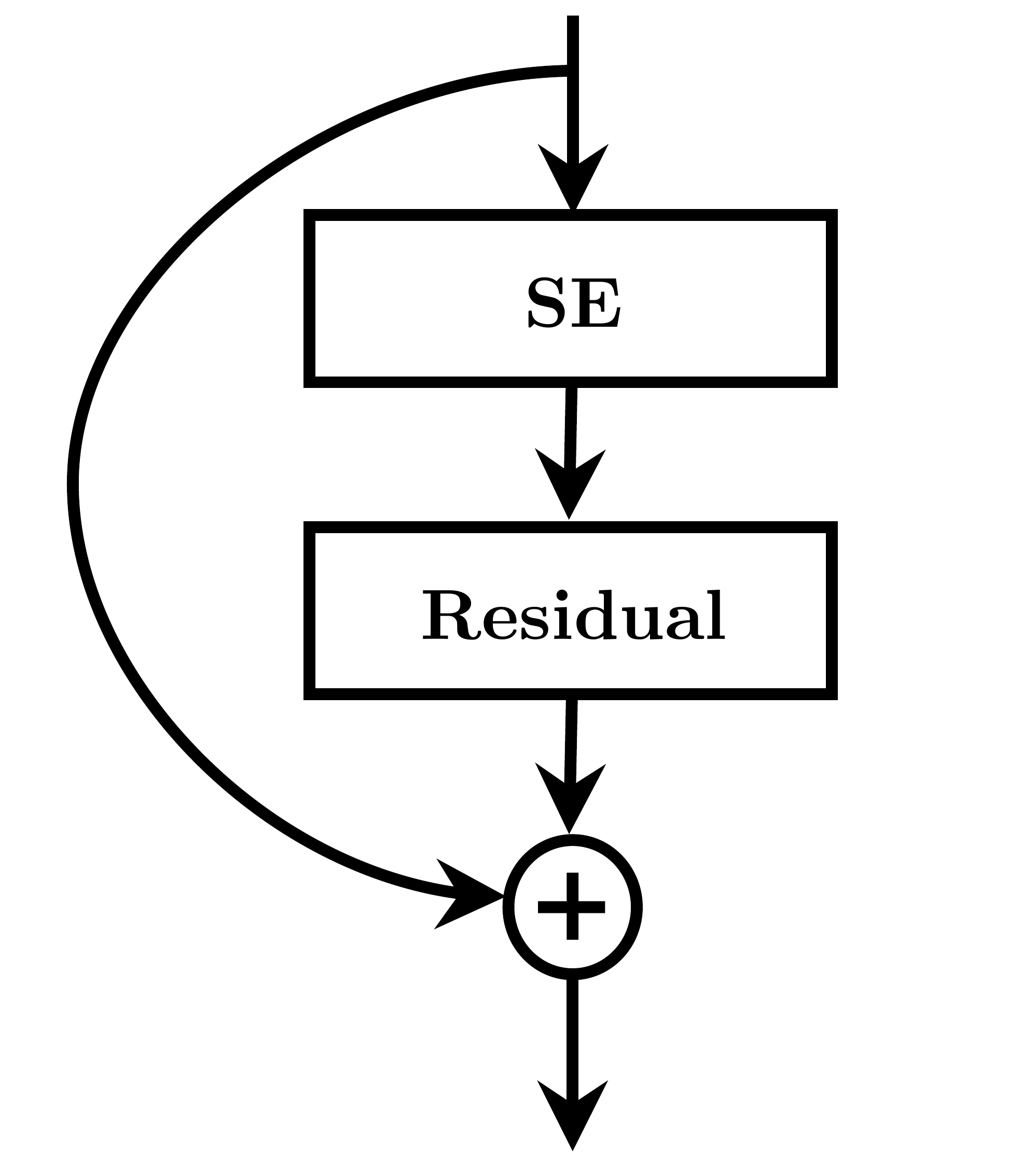}} \hspace{3mm}
	\subfigure[SE-POST block]{\includegraphics[width=.16 \textwidth]{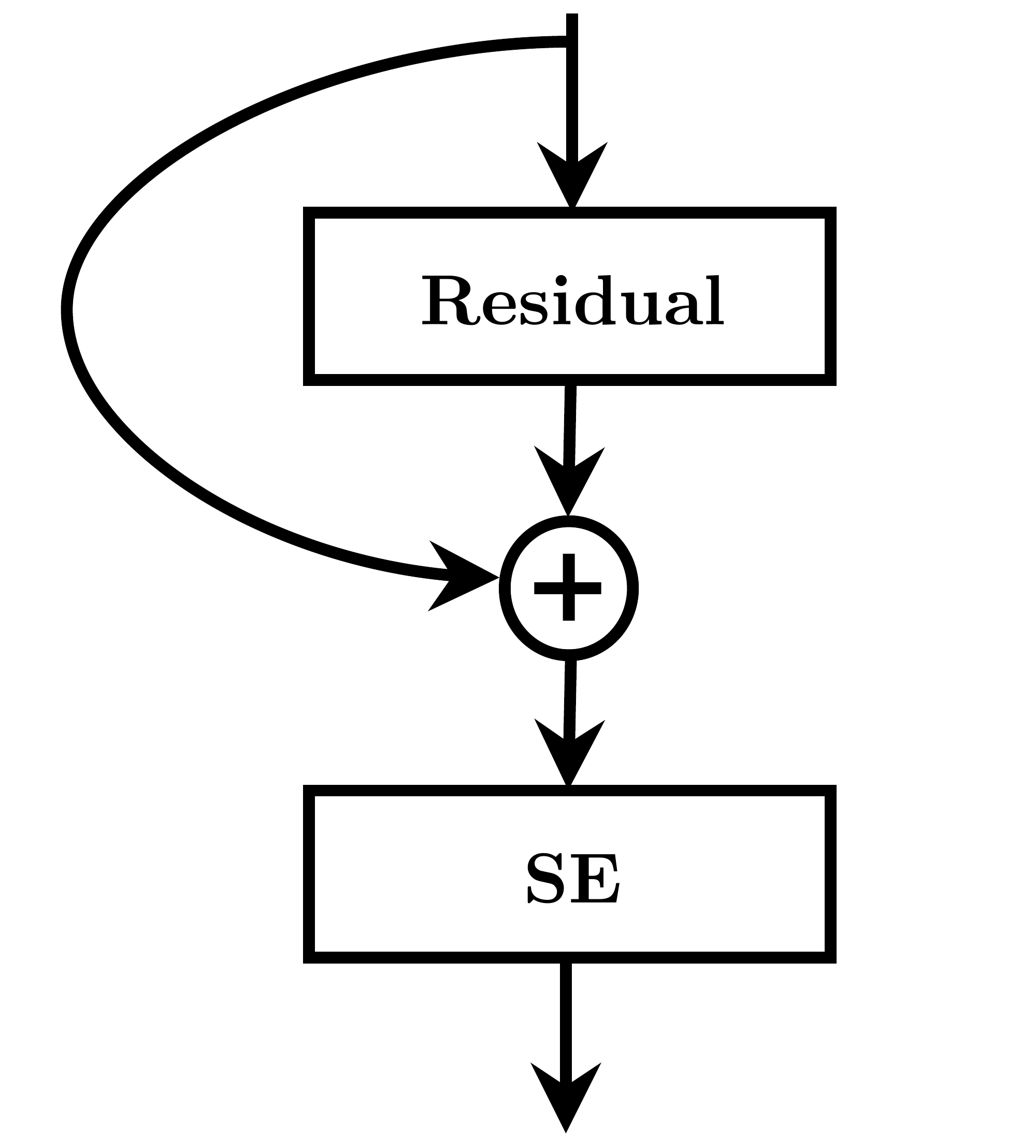}} \hspace{3mm}
   \subfigure[SE-Identity block]{\includegraphics[width=.16 \textwidth]{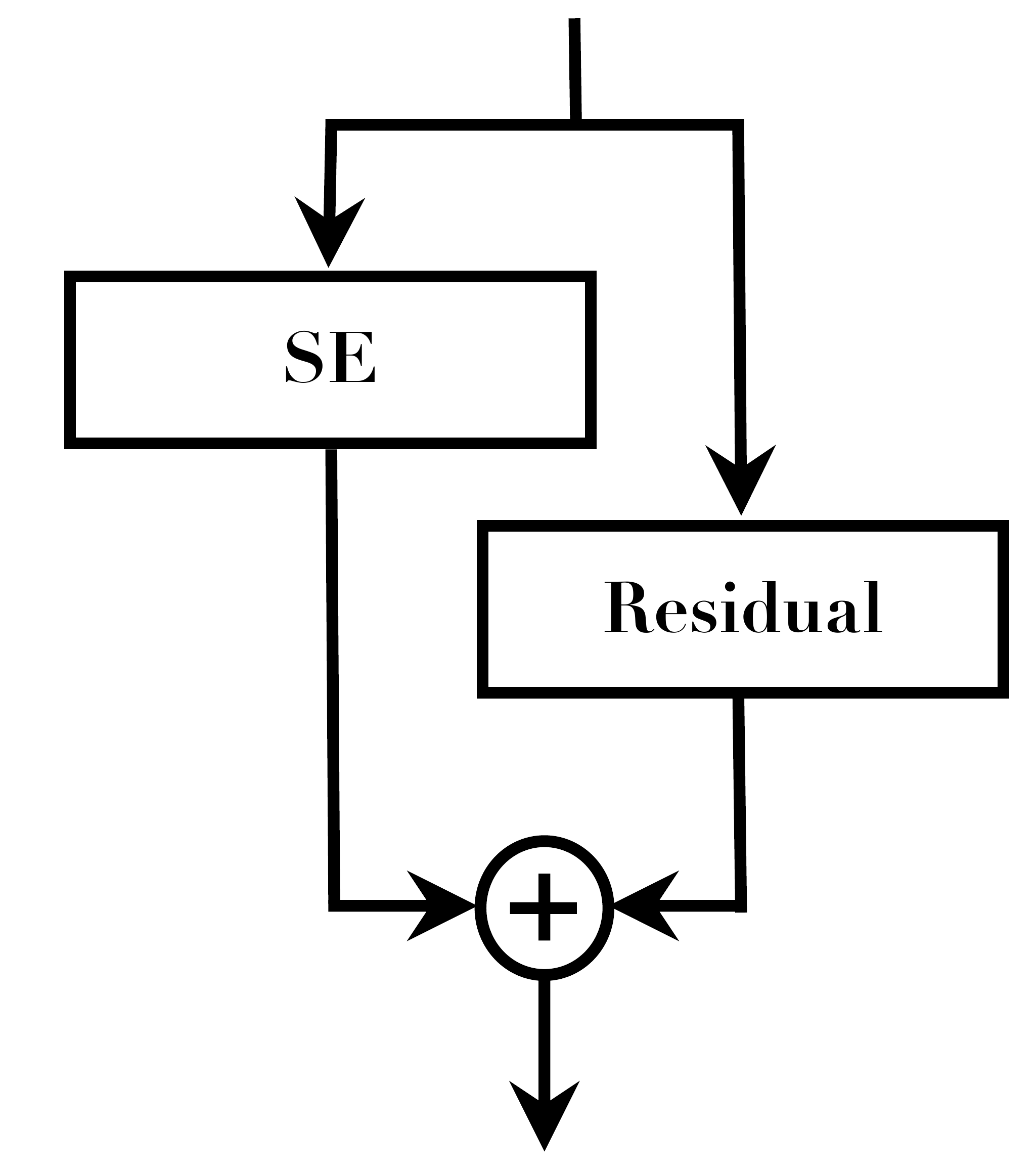}} 
\end{center}
\vspace{-1.3em}
\caption{SE block integration designs explored in the ablation study.}
\label{fig:ablation-designs}
\end{figure*}

Finally, we perform an ablation study to assess the influence of the location of the SE block when integrating it into existing architectures.  In addition to the proposed SE design, we consider three variants: (1) SE-PRE block, in which the SE block is moved before the residual unit; (2) SE-POST block, in which the SE unit is moved after the summation with the identity branch (after ReLU) and (3) SE-Identity block, in which the SE unit is placed on the identity connection in parallel to the residual unit.  These variants are illustrated in Figure~\ref{fig:ablation-designs} and the performance of each variant is reported in Table~\ref{tab:block-variants}.  We observe that the \mbox{SE-PRE}, \mbox{SE-Identity} and proposed SE block each perform similarly well, while usage of the \mbox{SE-POST} block leads to a drop in performance.  This experiment suggests that the performance improvements produced by SE units are fairly robust to their location, provided that they are applied prior to branch aggregation.
%--------------table----------------------------------------------

%------------------------------------- table --------------------------------------------
\begin{table}[t]
\renewcommand\arraystretch{1.1}
\small
\caption{Effect of different SE block integration strategies with ResNet-50 on ImageNet (error rates \%).}
\label{tab:block-variants}
\vspace{-1.7em}
\begin{center}{\scalebox{0.95}{
\begin{tabular}{l|p{1.3cm}<{\centering}|p{1.3cm}<{\centering}}
\hline
Design & top-1 err. & top-5 err.\\
\hline
SE       & 22.28 & 6.03 \\
SE-PRE  & 22.23  &  6.00   \\
SE-POST & 22.78  &  6.35   \\
SE-Identity   & 22.20  &  6.15  \\
\hline
\end{tabular}}}
\end{center}
\end{table}

\begin{table}[t]
\renewcommand\arraystretch{1.1}
\caption{Effect of integrating SE blocks at the 3x3 convolutional layer of each residual branch in ResNet-50 on ImageNet (error rates \%).}
\label{tab:effect_se3x3}
\vspace{-1.7em}
\begin{center}
\begin{tabular}{l|p{1.2cm}<{\centering}|p{1.2cm}<{\centering}|p{1.2cm} <{\centering}|p{1.2cm} <{\centering}}
\hline
Design & top-1 err. & top-5 err. & GFLOPs & Params\\
\hline
SE & $22.28$  &  $6.03$ & $3.87$ & $28.1$M\\
SE\_3$\times$3 & $22.48$ & $6.02$ & $3.86$ & $25.8$M\\
\hline
\end{tabular}
\end{center}
\end{table}

In the experiments above, each SE block was placed outside the structure of a residual unit. We also construct a variant of the design which moves the SE block inside the residual unit, placing it directly after the $3\times3$ convolutional layer. Since the $3\times3$ convolutional layer possesses fewer channels, the number of parameters introduced by the corresponding SE block is also reduced. The comparison in Table~\ref{tab:effect_se3x3} shows that the SE\_3$\times$3 variant achieves comparable classification accuracy with fewer parameters than the standard SE block.  Although it is beyond the scope of this work, we anticipate that further efficiency gains will be achievable by tailoring SE block usage for specific architectures.
\section{Role of SE blocks} \label{sec:interpretation}

Although the proposed SE block has been shown to improve network performance on multiple visual tasks, we would also like to understand the relative importance of the squeeze operation and how the excitation mechanism operates in practice. A rigorous theoretical analysis of the representations learned by deep neural networks remains challenging, we therefore take an empirical approach to examining the role played by the SE block with the goal of attaining at least a primitive understanding of its practical function.
%---------------table and figure ----------------------------
\begin{table}[t]
\renewcommand\arraystretch{1.1}
\caption{Effect of Squeeze operator on ImageNet (error rates \%).}
\label{tab:effect_squeeze}
\vspace{-1.7em}
\small
\begin{center}{\scalebox{0.96}{
\begin{tabular}{l|p{1.3cm}<{\centering}|p{1.3cm}<{\centering}|p{1.2cm} <{\centering}|p{1.2cm} <{\centering}}
\hline
& top-1 err. & top-5 err. & GFLOPs & Params\\
\hline
ResNet-50 &   $23.30$ &   $6.55$ & $3.86$ & $25.6$M\\
NoSqueeze & $22.93$ & $6.39$ & $4.27$ & $28.1$M\\
SE & $\mathbf{22.28}$  &  $\mathbf{6.03}$ & $3.87$ & $28.1$M\\
\hline
\end{tabular}}}
\end{center}
\end{table}

\begin{figure*}[t]
\small
\begin{center}
\subfigure[\texttt{SE\_2\_3}]{\includegraphics[trim={1cm 7cm 1cm 7cm},clip,width=0.28\textwidth,height=0.15\textheight]{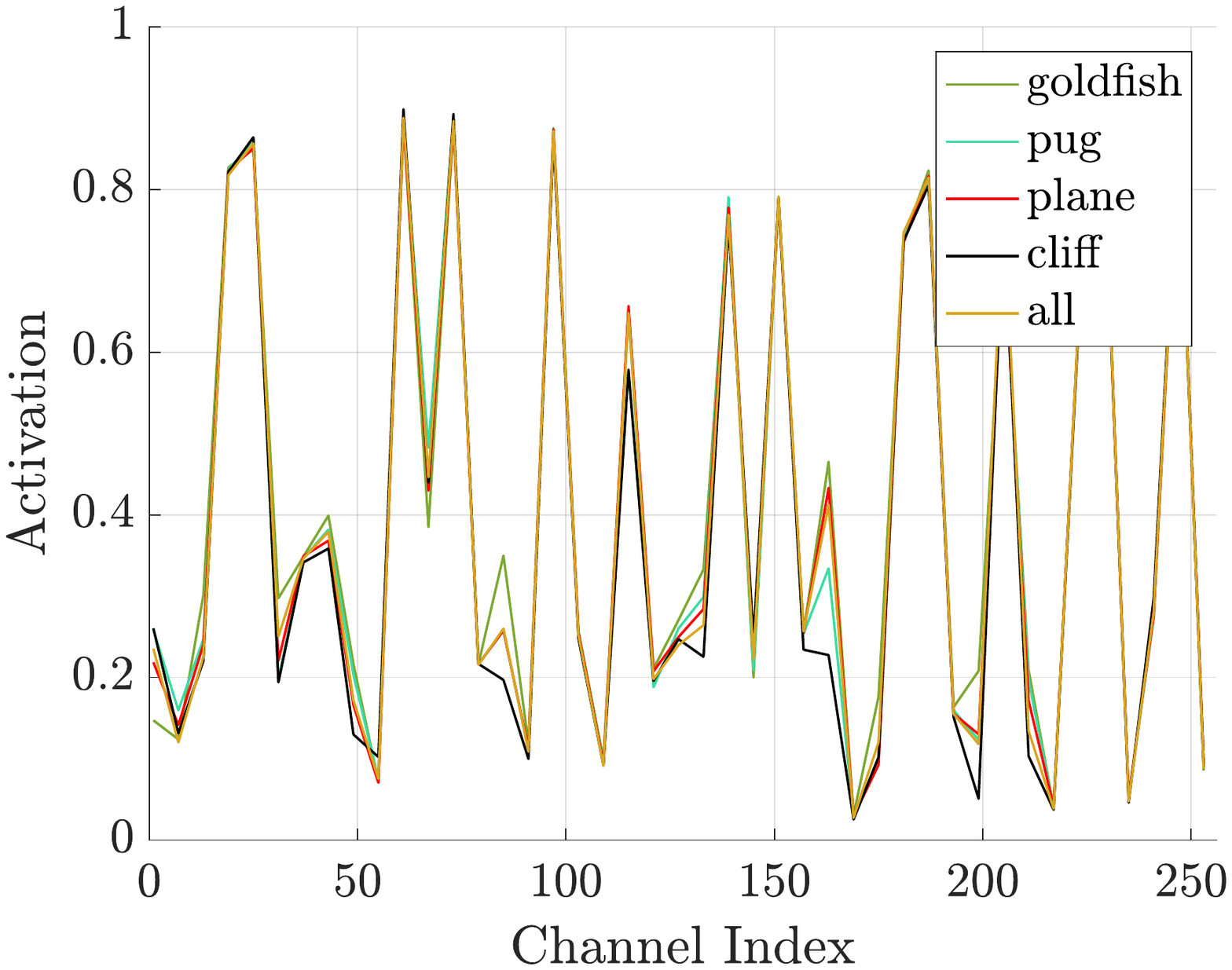}}\hspace{0.2cm}
\subfigure[\texttt{SE\_3\_4}]{\includegraphics[trim={1cm 7cm 1cm 7cm},clip,width=0.28\textwidth,height=0.15\textheight]{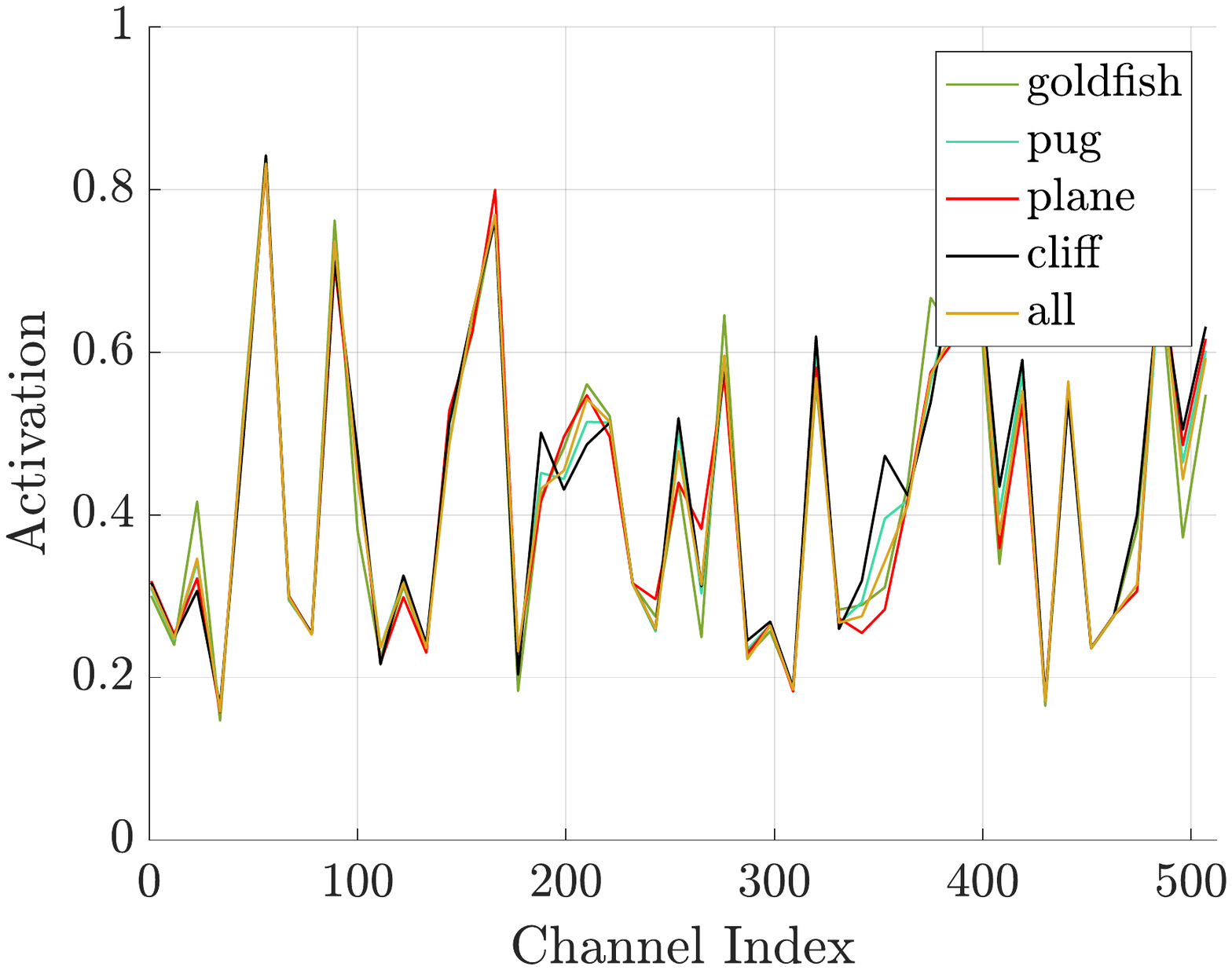}}
    \hspace{0.2cm}
\subfigure[\texttt{SE\_4\_6}]{\includegraphics[trim={1cm 7cm 1cm 7cm},clip,width=0.28\textwidth,height=0.15\textheight]{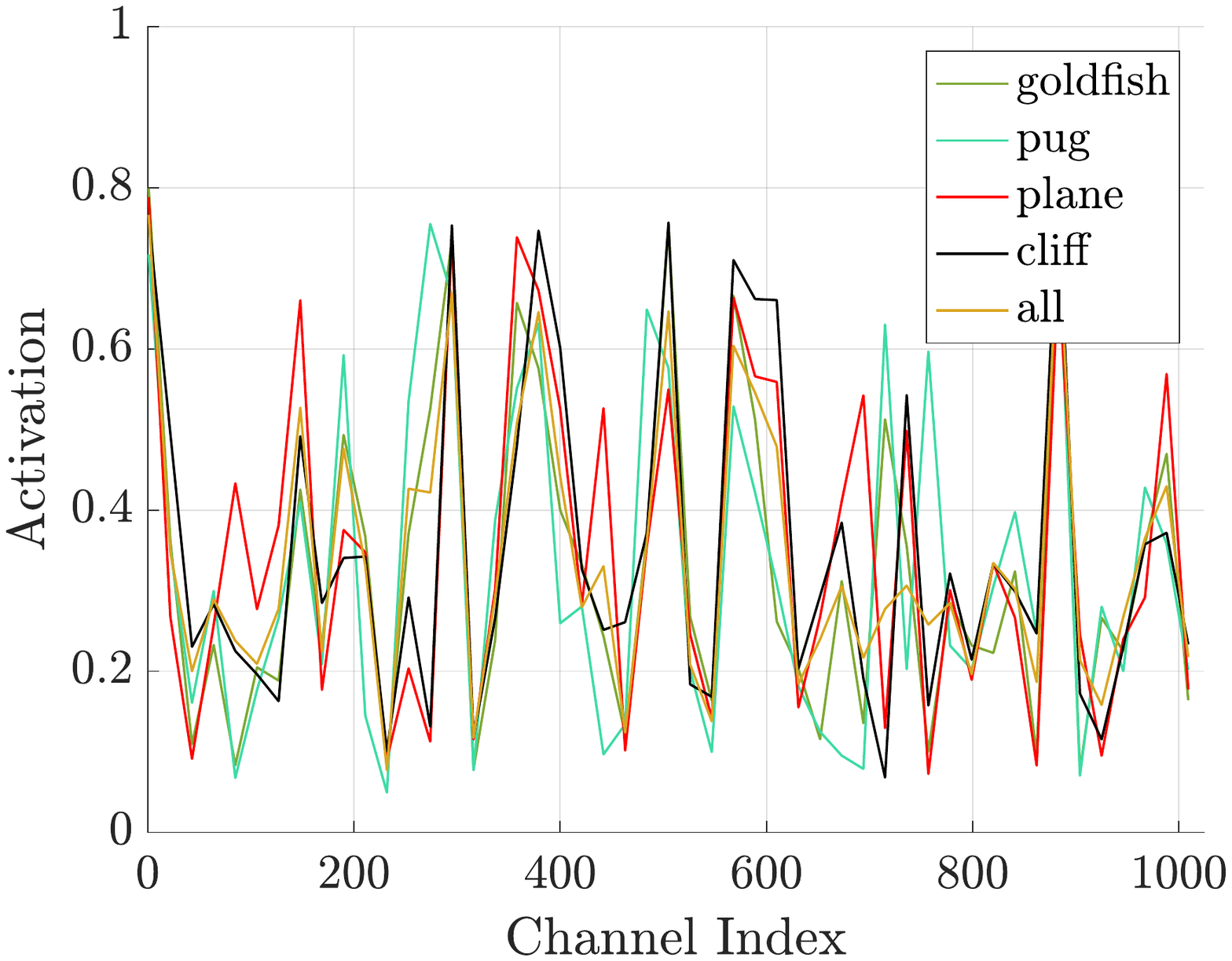}}
 \hspace{0.2cm} 
\subfigure[\texttt{SE\_5\_1}]{\includegraphics[trim={1cm 7cm 1cm 7cm},clip,width=0.28\textwidth,height=0.15\textheight]{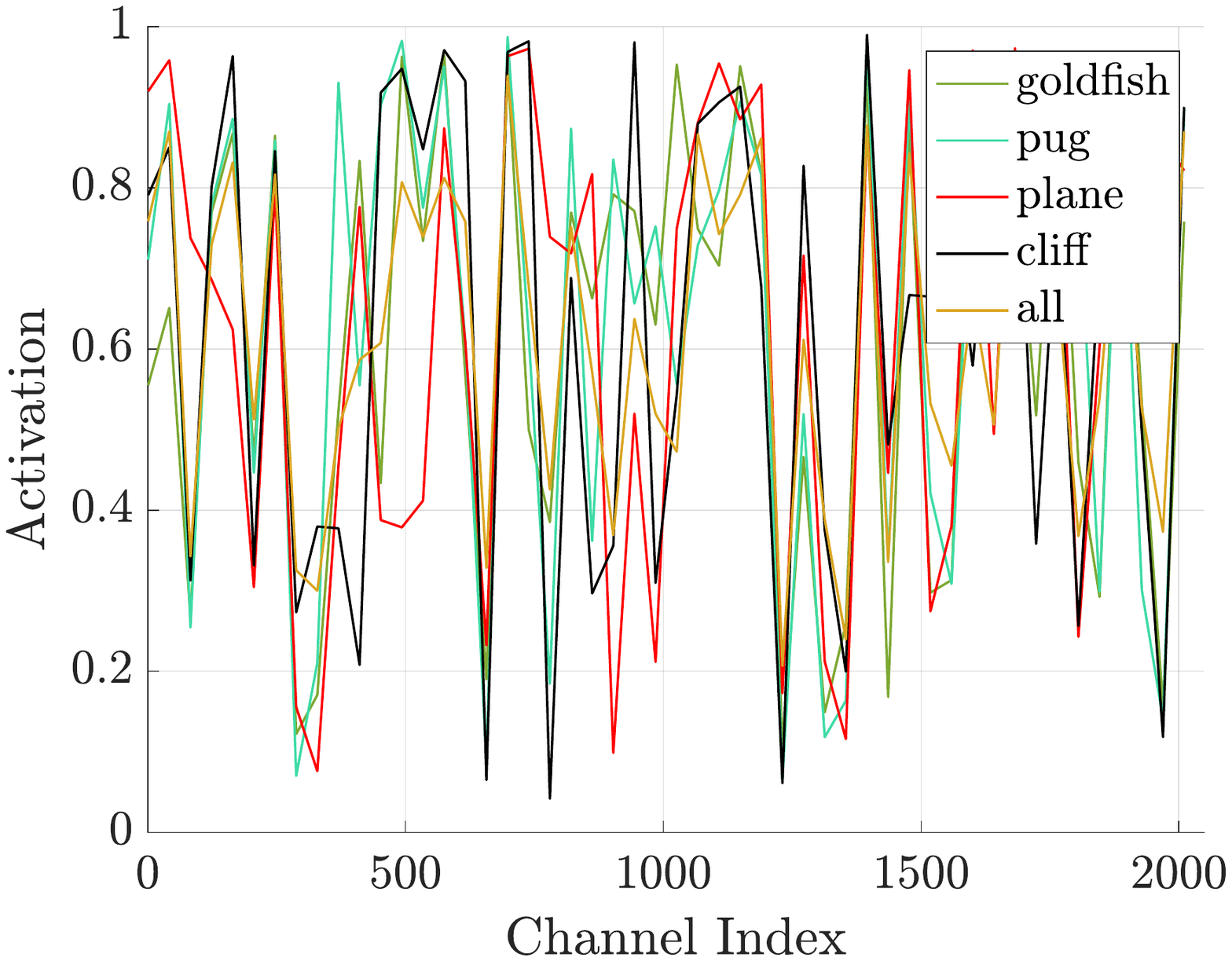}}
 \hspace{0.2cm} 
\subfigure[\texttt{SE\_5\_2}]{\includegraphics[trim={1cm 7cm 1cm 7cm},clip,width=0.28\textwidth,height=0.15\textheight]{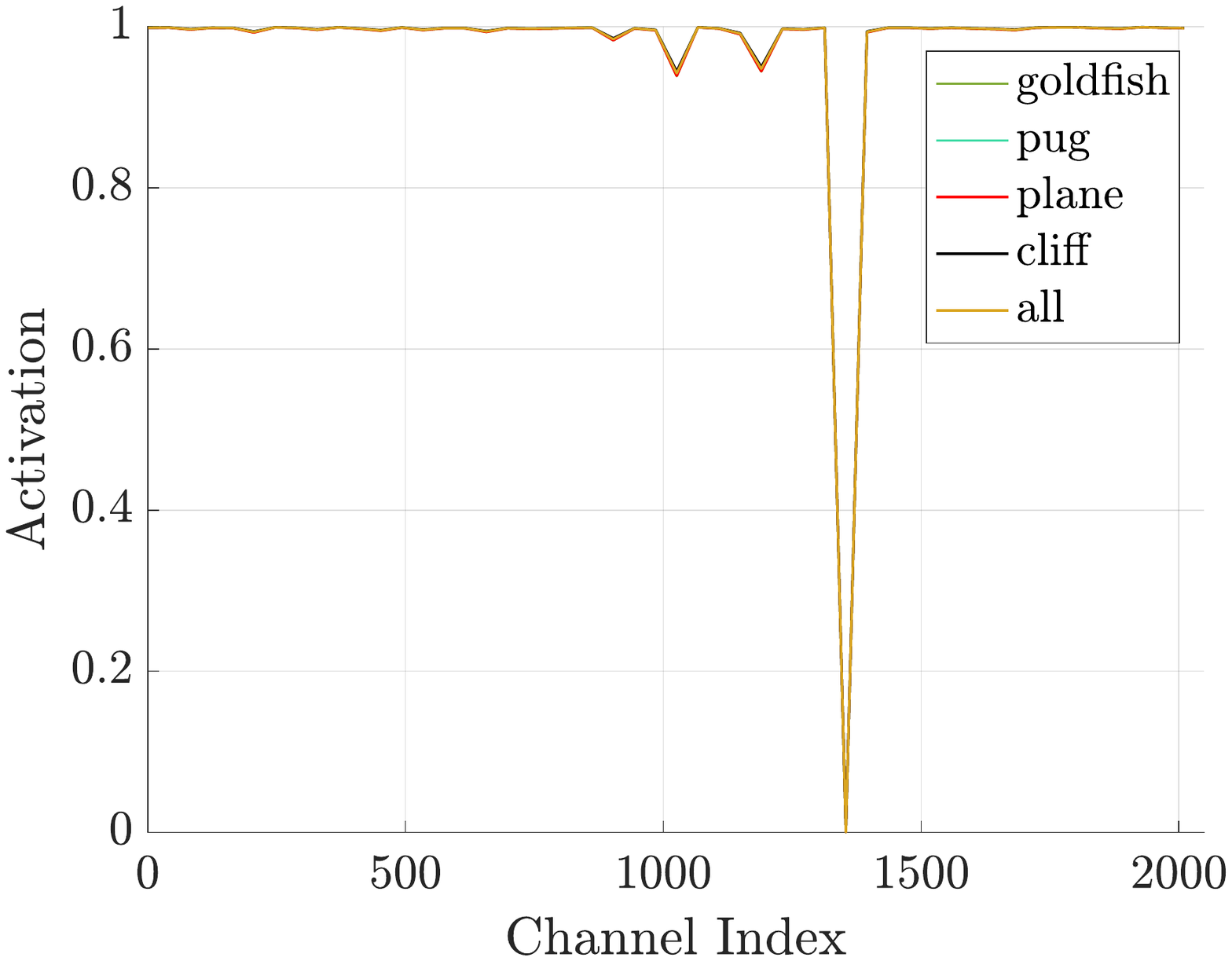}}
 \hspace{0.2cm}
\subfigure[\texttt{SE\_5\_3}]{\includegraphics[trim={1cm 7cm 1cm 7cm},clip,width=0.28\textwidth,height=0.15\textheight]{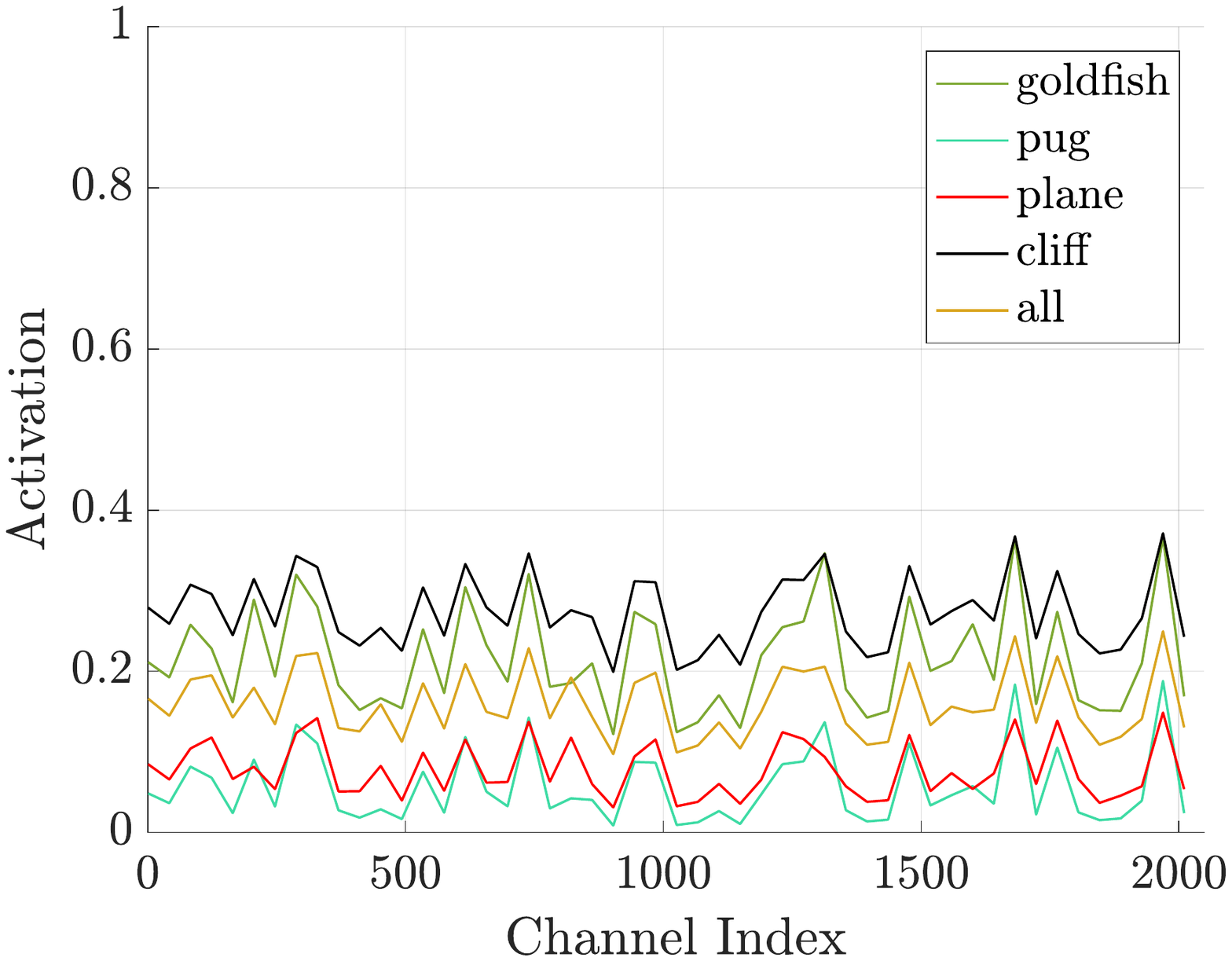}}
\end{center}
\vspace{-0.5cm}
\caption{Activations induced by the Excitation operator at different depths in the SE-ResNet-50 on ImageNet. Each set of activations is named according to the following scheme: \texttt{SE\_stageID\_blockID}.  With the exception of the unusual behaviour at \texttt{SE\_5\_2}, the activations become increasingly class-specific with increasing depth.}
\label{fig:class-activations}
\end{figure*}

\begin{figure*}[t]
\small
\begin{center}
\subfigure[\texttt{SE\_2\_3}]{\includegraphics[trim={1cm 7cm 1cm 7cm},clip,width=0.28\textwidth,height=0.15\textheight]{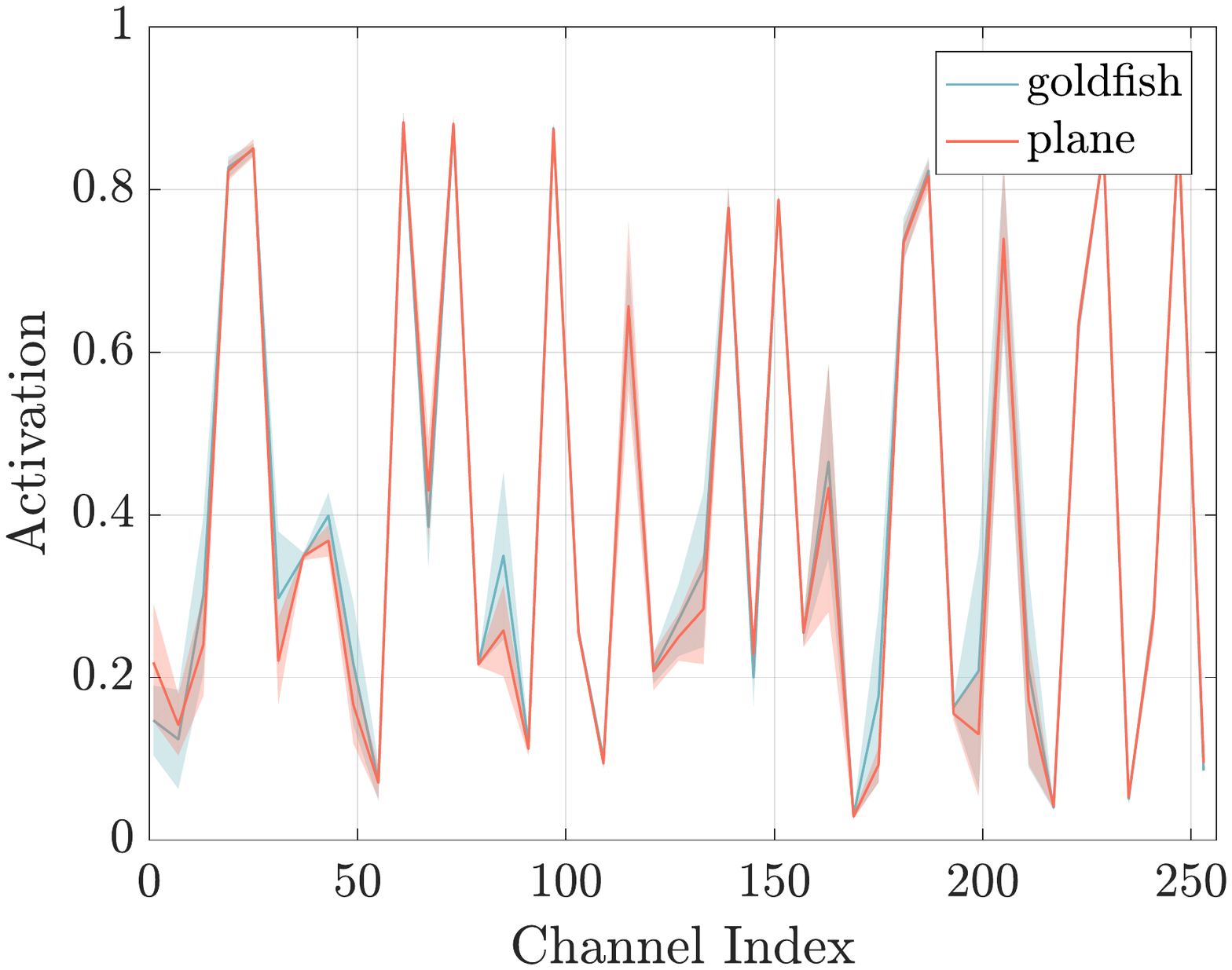}} \hspace{0.2cm}
\subfigure[\texttt{SE\_3\_4}]{\includegraphics[trim={1cm 7cm 1cm 7cm},clip,width=0.28\textwidth,height=0.15\textheight]{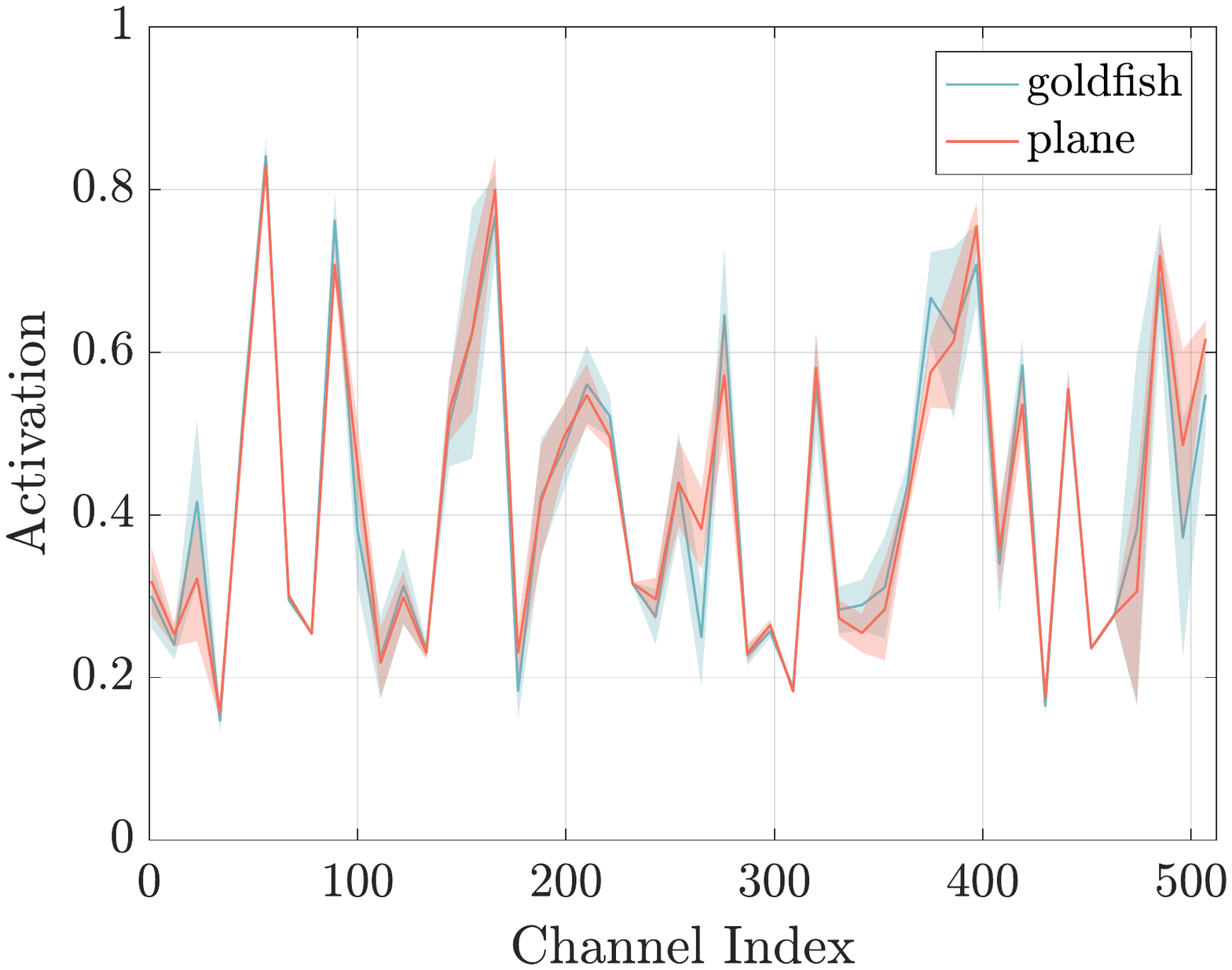}} \hspace{0.2cm}
\subfigure[\texttt{SE\_4\_6}]{\includegraphics[trim={1cm 7cm 1cm 7cm},clip,width=0.28\textwidth,height=0.15\textheight]{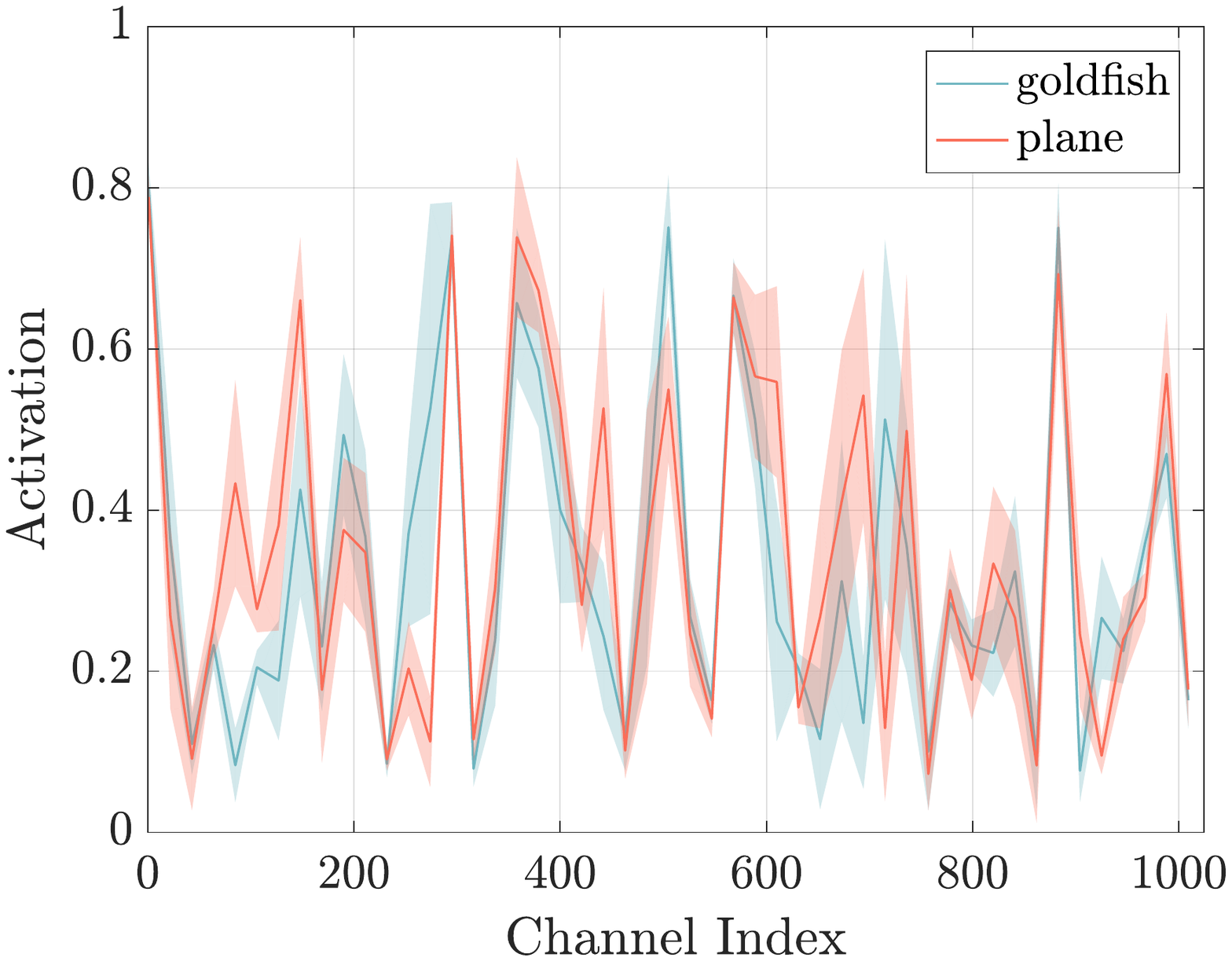}}  \hspace{0.2cm}
\subfigure[\texttt{SE\_5\_1}]{\includegraphics[trim={1cm 7cm 1cm 7cm},clip,width=0.28\textwidth,height=0.15\textheight]{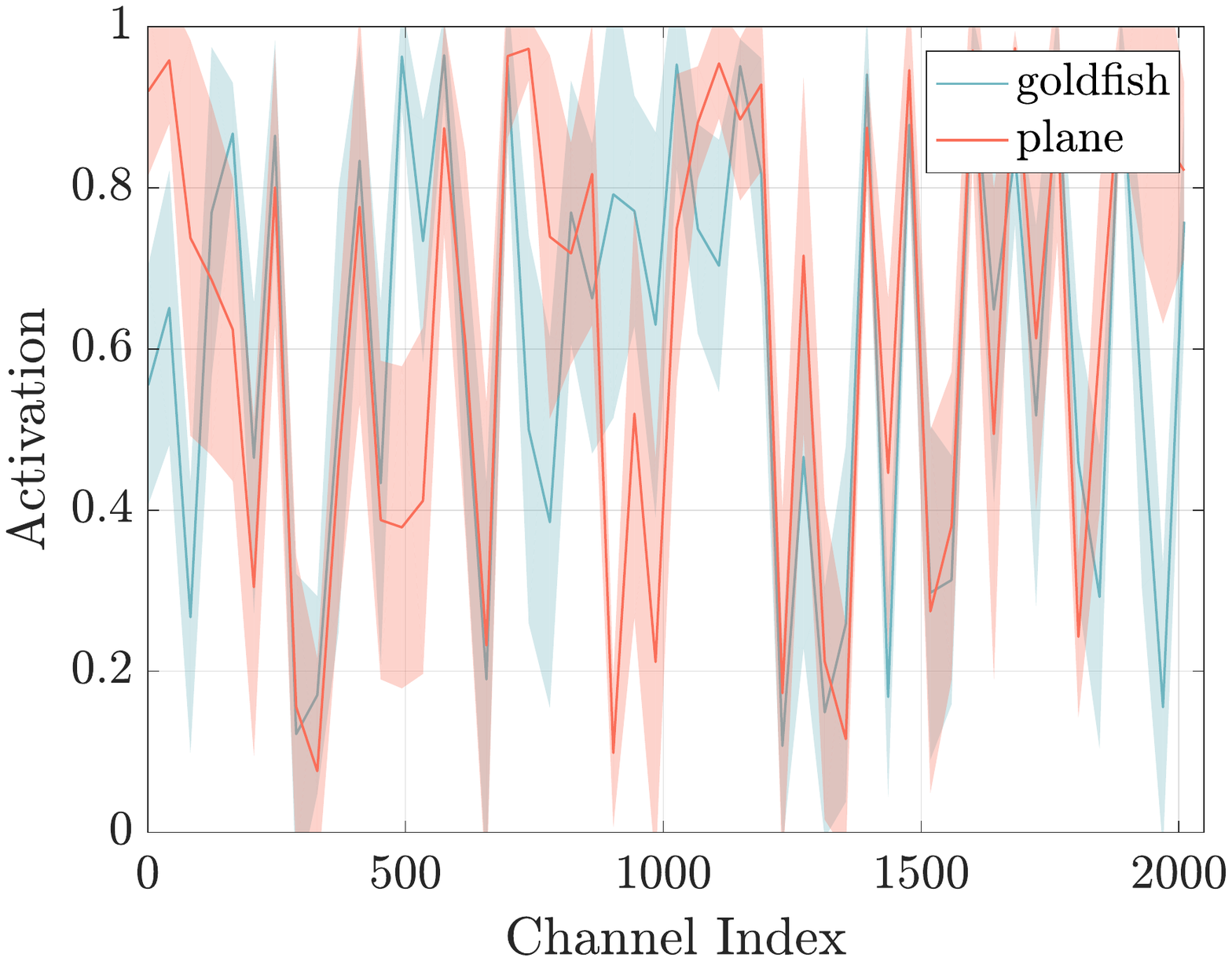}}  \hspace{0.2cm}
\subfigure[\texttt{SE\_5\_2}]{\includegraphics[trim={1cm 7cm 1cm 7cm},clip,width=0.28\textwidth,height=0.15\textheight]{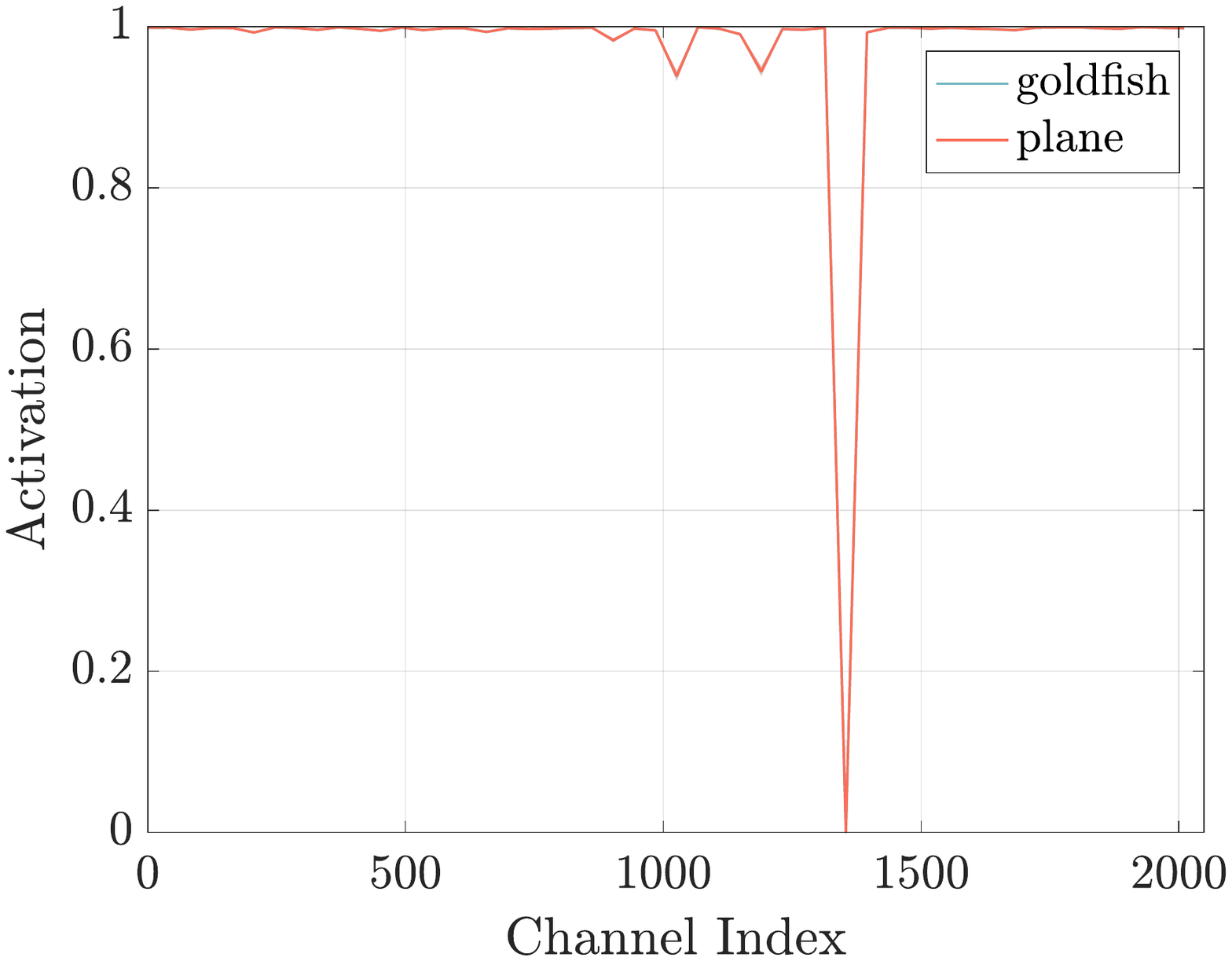}} \hspace{0.2cm}
\subfigure[\texttt{SE\_5\_3}]{\includegraphics[trim={1cm 7cm 1cm 7cm},clip,width=0.28\textwidth,height=0.15\textheight]{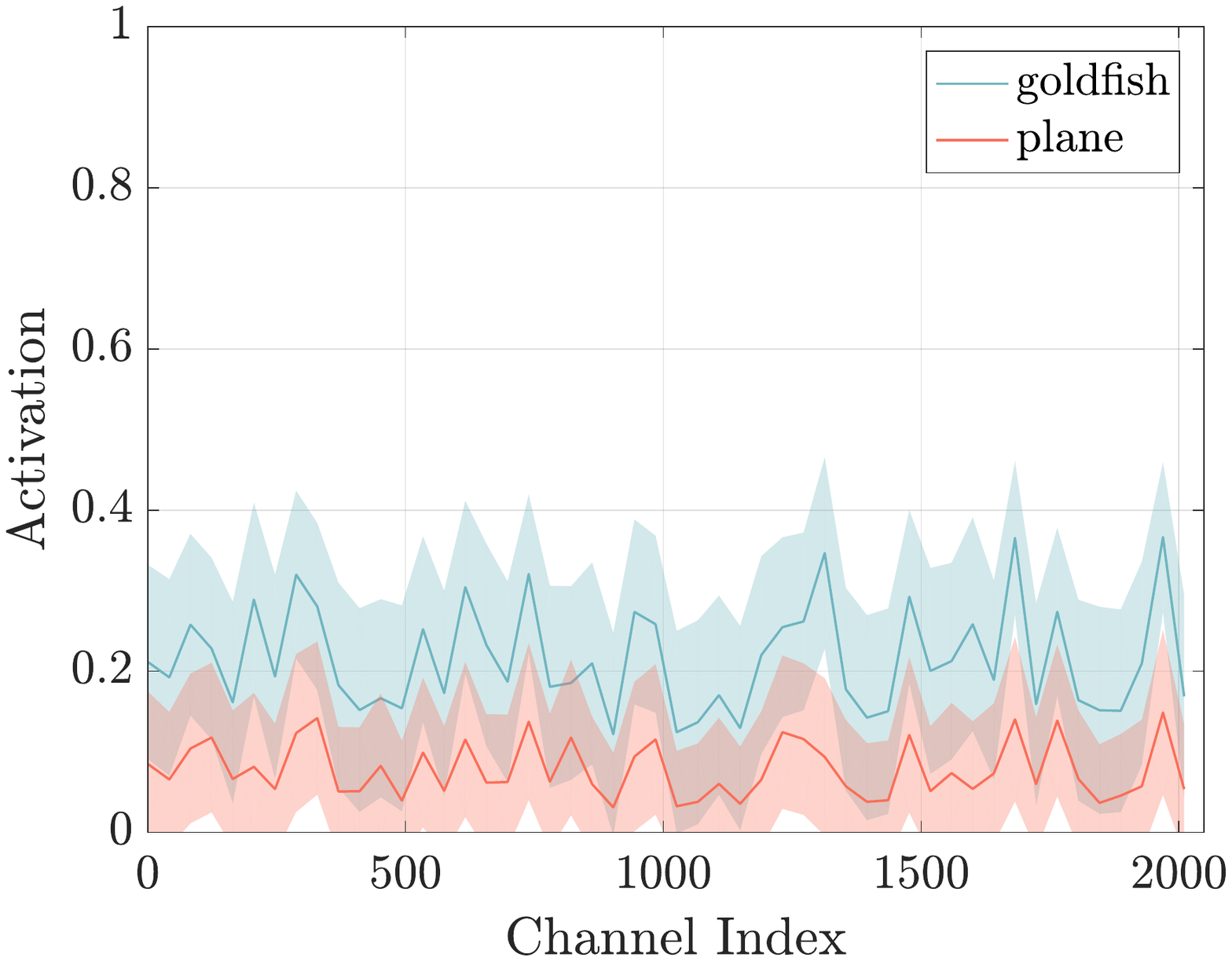}} \hspace{0.2cm}
\end{center}
\vspace{-1.5em}
\caption{Activations induced by {\it Excitation} in the different modules of SE-ResNet-50 on image samples from the goldfish and plane classes of ImageNet. The module is named  ``\texttt{SE\_stageID\_blockID}''.}
\label{fig:instance-activations}
\end{figure*}

\subsection{Effect of Squeeze}\label{subsection:effect-of-squeeze}
To assess whether the global embedding produced by the squeeze operation plays an important role in performance, we experiment with a variant of the SE block that adds an equal number of parameters, but does not perform global average pooling. Specifically, we remove the pooling operation and replace the two FC layers with corresponding $1\times 1$ convolutions with identical channel dimensions in the excitation operator, namely \textit{NoSqueeze}, where the excitation output maintains the spatial dimensions as input. In contrast to the SE block, these point-wise convolutions can only remap the channels as a function of the output of a local operator. While in practice, the later layers of a deep network will typically possess a (theoretical) global receptive field, global embeddings are no longer directly accessible throughout the network in the \textit{NoSqueeze} variant. The accuracy and computational complexity of both models are compared to a standard ResNet-50 model in Table~\ref{tab:effect_squeeze}. We observe that the use of global information has a significant influence on the model performance, underlining the importance of the squeeze operation.  Moreover, in comparison to the \textit{NoSqueeze} design, the SE block allows this global information to be used in a computationally parsimonious manner.

\subsection{Role of Excitation} \label{subsection:role-of-excitation}
To provide a clearer picture of the function of the excitation operator in SE blocks, in this section we study example activations from the SE-ResNet-50 model and examine their distribution with respect to different classes and different input images at various depths in the network.  In particular, we would like to understand how excitations vary across images of different classes, and across images within a class.

We first consider the distribution of excitations for different classes. Specifically, we sample four classes from the ImageNet dataset that exhibit semantic and appearance diversity, namely \textit{goldfish}, \textit{pug}, \textit{plane} and \textit{cliff} (example images from these classes are shown in Appendix). We then draw fifty samples for each class from the validation set and compute the average activations for fifty uniformly sampled channels in the last SE block of each stage (immediately prior to downsampling) and plot their distribution in Fig.~\ref{fig:class-activations}. For reference, we also plot the distribution of the mean activations across all of the $1000$ classes.

We make the following three observations about the role of the {\it excitation} operation. First, the distribution across different classes is very similar at the earlier layers of the network, e.g. SE\_2\_3.  This suggests that the importance of feature channels is likely to be shared by different classes in the early stages.  The second observation is that at greater depth, the value of each channel becomes much more \mbox{class-specific} as different classes exhibit different preferences to the discriminative value of features, e.g. SE\_4\_6 and SE\_5\_1.  These observations are consistent with findings in previous work \cite{lee_icml2009cdbn, yosinki_nips2014transfer}, namely that earlier layer features are typically more general (e.g. class agnostic in the context of the classification task) while later layer features exhibit greater levels of specificity \cite{morcos_iclr2018importance}.

Next, we observe a somewhat different phenomena in the last stage of the network. SE\_5\_2 exhibits an interesting tendency towards a saturated state in which most of the activations are close to one. At the point at which all activations take the value one, an SE block reduces to the identity operator. At the end of the network in the SE\_5\_3 (which is immediately followed by global pooling prior before classifiers), a similar pattern emerges over different classes, up to a modest change in scale (which could be tuned by the classifiers).  This suggests that SE\_5\_2 and SE\_5\_3 are less important than previous blocks in providing recalibration to the network.  This finding is consistent with the result of the empirical investigation in Section~\ref{sec:modelcapacity} which demonstrated that the additional parameter count could be significantly reduced by removing the SE blocks for the last stage with only a marginal loss of performance.

Finally, we show the mean and standard deviations of the activations for image instances within the same class for two sample classes (\textit{goldfish} and \textit{plane}) in Fig.~\ref{fig:instance-activations}.  We observe a trend consistent with the inter-class visualisation, indicating that the dynamic behaviour of SE blocks varies over both classes and instances within a class. Particularly in the later layers of the network where there is considerable diversity of representation within a single class, the network learns to take advantage of feature recalibration to improve its discriminative performance \cite{hu2018GE}. In summary, SE blocks produce instance-specific responses which nevertheless function to support the increasingly class-specific needs of the model at different layers in the architecture.
\section{Conclusion}

In this paper we proposed the SE block, an architectural unit designed to improve the representational power of a network by enabling it to perform dynamic channel-wise feature recalibration. A wide range of experiments show the effectiveness of SENets, which achieve state-of-the-art performance across multiple datasets and tasks. In addition, SE blocks shed some light on the inability of previous architectures to adequately model channel-wise feature dependencies. We hope this insight may prove useful for other tasks requiring strong discriminative features. Finally, the feature importance values produced by SE blocks may be of use for other tasks such as network pruning for model compression.

\ifCLASSOPTIONcompsoc
  \section*{Acknowledgments}
\else
  \section*{Acknowledgment}
\fi

The authors would like to thank Chao Li and Guangyuan Wang from Momenta for their contributions in the training system optimisation and experiments on CIFAR dataset. We would also like to thank Andrew Zisserman, Aravindh Mahendran and Andrea Vedaldi for many helpful discussions. The work is supported in part by NSFC Grants (61632003, 61620106003, 61672502, 61571439), National Key R\&D Program of China (2017YFB1002701), and Macao FDCT Grant (068/2015/A2). Samuel Albanie is supported by EPSRC AIMS CDT EP/L015897/1.

\ifCLASSOPTIONcaptionsoff
  \newpage
\fi

%\appendices 
\section*{Appendix: Details of SENet-154} \label{app:senet154}

SENet-154 is constructed by incorporating SE blocks into a modified version of the 64$\times$4d ResNeXt-152 which extends the original ResNeXt-101 \cite{xie_cvpr2017resnext} by adopting the block stacking strategy of ResNet-152 \cite{he_cvpr2016resnet}. Further differences to the design and training of this model (beyond the use of SE blocks) are as follows:
(a) The number of the first $1\times1$ convolutional channels for each bottleneck building block was halved to reduce the computational cost of the model with a minimal decrease in performance.
(b) The first $7\times7$ convolutional layer was replaced with three consecutive $3\times3$ convolutional layers.
(c) The $1\times1$ down-sampling projection with stride-$2$ convolution was replaced with a $3\times3$ stride-$2$ convolution to preserve information.
(d) A dropout layer (with a dropout ratio of $0.2$) was inserted before the classification layer to reduce overfitting.
(e) Label-smoothing regularisation (as introduced in \cite{szegedy_cvpr2016inceptionv3}) was used during training.
(f) The parameters of all BN layers were frozen for the last few training epochs to ensure consistency between training and testing.
(g) Training was performed with 8 servers (64 GPUs) in parallel to enable large batch sizes (2048).  The initial learning rate was set to $1.0$.

\begin{figure}[t]
\begin{center}
\subfigure[goldfish]{\includegraphics[height=16mm]{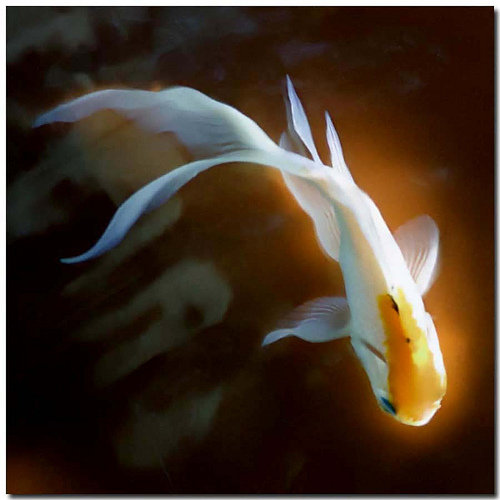}}  \hspace{0.5mm}
\subfigure[pug]{\includegraphics[height=16mm]{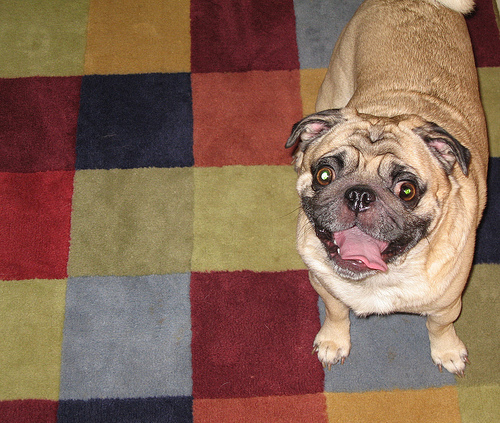}} 
    \hspace{0.5mm}
\subfigure[plane]{\includegraphics[height=16mm]{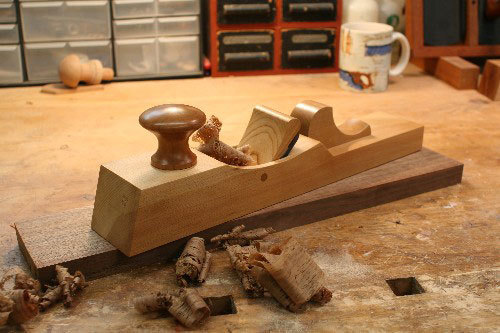}} 
    \hspace{0.5mm}
\subfigure[cliff]{\includegraphics[height=16mm]{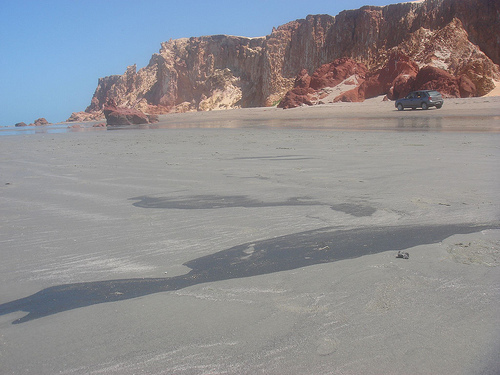}}
    \hspace{0.5mm}
\end{center}
\vspace{-1.5em}
\caption{Sample images from the four classes of ImageNet used in the experiments described in Sec.~7.2.%~\ref{subsection:role-of-excitation}.
}
\label{fig:sample-ims}
\end{figure}



\bibliographystyle{IEEEtran}
\bibliography{references}

\end{document}